\documentclass[conference]{IEEEtran}
\IEEEoverridecommandlockouts
\usepackage{cite}
\usepackage{amsmath,amssymb,amsfonts}
\usepackage{algorithmic}
\usepackage{graphicx}
\usepackage{textcomp}

\usepackage{xcolor}
\usepackage{booktabs}
\usepackage{amsmath}
\usepackage{bbding}
\usepackage{hyperref}
\usepackage{bm}

\newcommand{\tabincell}[2]{\begin{tabular}{@{}#1@{}}#2\end{tabular}}
\def\BibTeX{{\rm B\kern-.05em{\sc i\kern-.025em b}\kern-.08em
    T\kern-.1667em\lower.7ex\hbox{E}\kern-.125emX}}

\begin{document}
\title{ A Secure and Efficient Multi-Object Grasping Detection Approach for Robotic Arms}

 \author{\IEEEauthorblockN{
 Hui Wang\textsuperscript{1,2},
 Jieren Cheng\textsuperscript{1,}\IEEEauthorrefmark{1}\thanks{*Corresponding author: Jieren Cheng},
 Yichen Xu\textsuperscript{1,2},
 Sirui Ni\textsuperscript{1},
 Zaijia Yang\textsuperscript{1},
 Jiangpeng Li\textsuperscript{1},
}

 \IEEEauthorblockA{\textsuperscript{1}School of Computer Science and Technology and \\ School of Information and Communication Engineering, Hainan University, Haikou, China}
 \IEEEauthorblockA{\textsuperscript{2}RobAI-Lab, Hainan University, Haikou, China}
 \IEEEauthorblockA{\{ whmio0115 ,star ,siruini , 20213002625 ,xquery\}@hainanu.edu.cn}
 }

\maketitle

\begin{abstract}
\par Robot grasping is one of the most important abilities of modern intelligent robots, especially industrial robots. However, most of the existing robot arm's grasp detection work is highly dependent on their edge computing ability, and the safety problems in the process of grasp detection are not considered enough. In this paper, we propose a new robotic arm grasping detection model with an edge-cloud collaboration method. With the scheme of multi-object multi-grasp, our model improves the mission success ratio of grasping. The model can not only complete the compression of full-resolution images but also achieve image compression at a limited bit rate. The image compression ratio reaches 2.03\%, the structural difference value is higher than 0.91, and our average detection speed reaches 13.62fps.  Furthermore, we have packaged our model as a functional package of the ROS operating system, which can be easily used in actual robotic arm operations. Our solution can be fully applied to other work of robots to promote the development of the field of robotics.
\end{abstract}

\section{Introduction}
Grasping ability is one of the most important abilities of modern intelligent robots, especially for industrial robots, which will bring great power to society\cite{sanchez2018robotic}.
As the most common basic action of robots in work, robotic autonomous grasping has great application prospects. Because of its significance, robotic autonomous grasping has been studied for a long time. Recently, robot grasping has made rapid progress due to the rapid development of deep learning. 
There are many tasks in robot grasping, including object localization, pose estimation, grasp detection, motion planning, etc. Among these tasks, grasp detection is a key task in the computer vision and robotics discipline and has been the subject of considerable research.
\par However, there are still numerous challenges to this task. On the one hand, the algorithm requires hardware computing power. With the widespread use of deep learning algorithms in grasp detection, deep learning models are deployed directly at the edge (robotic arms). 
And the hardware computing power is often not well executed, leading to delays and errors in data processing and grasp configuration. 
At present, most of the robotic arm's grasp detection work is calculated directly at the edge, only with the help of local computing power. 
This leads to the low efficiency of image detection, and can not meet the requirements of automatic grasp.
On the other hand, security issues in the process of grasp detection are often ignored, leading to the leakage of critical information. In recent years, there are also some studies that try to use cloud computing to solve the problem of insufficient local computing power. 
They upload the image data directly to the cloud (or fog), and with the help of the cloud's powerful computing power, this way greatly improves the efficiency of grasping. 
However, the direct transmission of data may lead to the problem of privacy leakage, while the transmission of real-time RGB images is often a major challenge for network bandwidth.

\begin{figure}[htbp]
\centerline{\includegraphics[height=5cm]{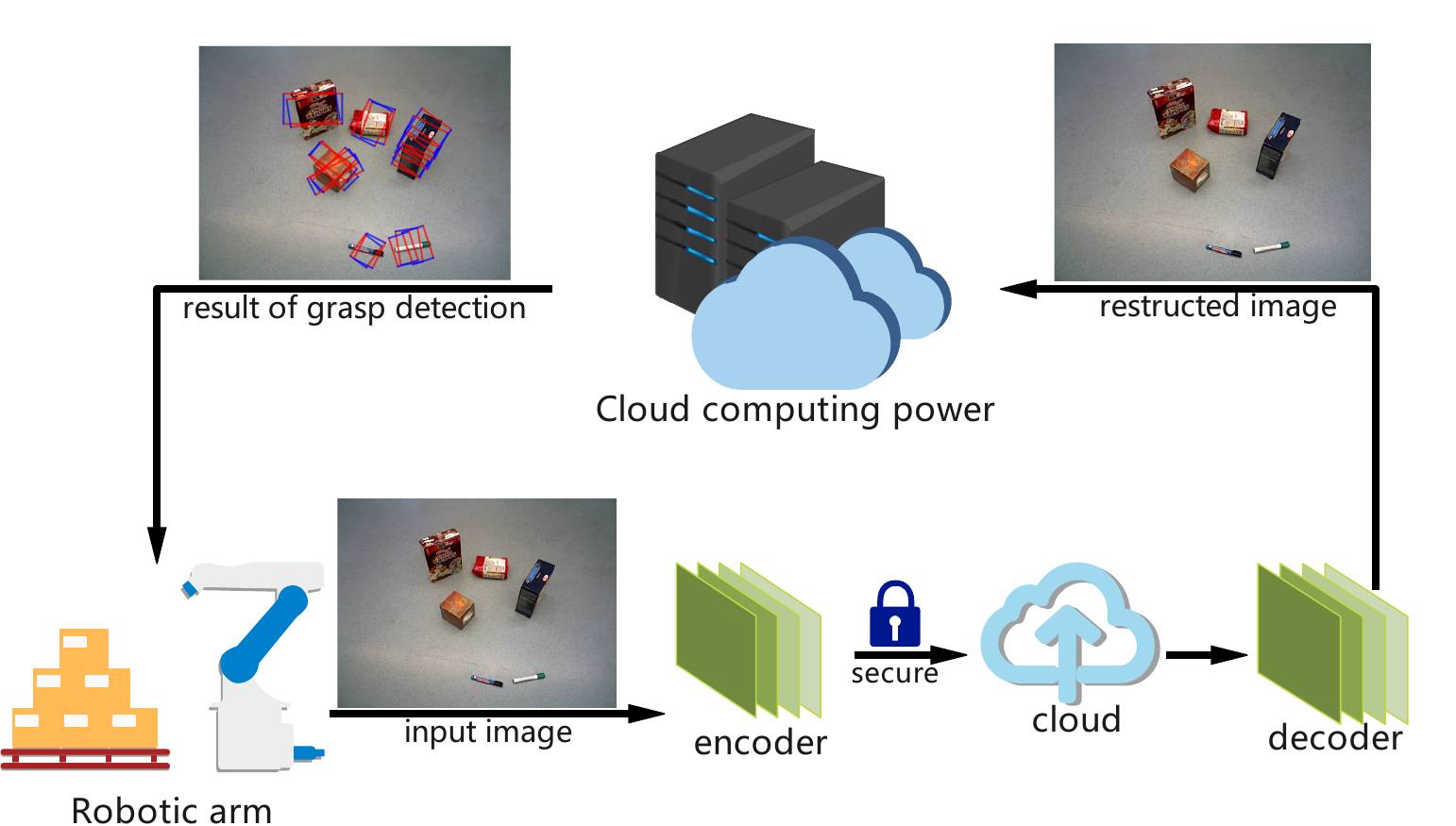}}
\caption{The figure shows how the robot arm unloads the local grasp detection task to the cloud. Our model realizes secure and high-fidelity transmission through this encoder-decoder structure. The image is collected locally, and transmitted to the cloud after being compressed. The reconstructed image will be obtained by the decoder in the cloud. Use cloud computing capabilities to assist in grasp detection and return the results to the robot arm.}
\label{fig1}
\end{figure}

\par In this work, we propose a robotic arm grasping detection model with an edge-cloud collaboration method. Figure 1 shows the execution flow of our technology model. We use an encoder to compress the images grasped by the camera locally and upload them to the cloud. 
The uploaded encoded information does not occupy local computing resources, and since it occupies less bandwidth and requires less network configuration, it is more suitable for real scenarios' deployment. 
In the cloud, our model reconstructs the image by a corresponding decoder, after which it performs a two-stage multi-object grasp detection and returns the obtained grasp configuration to the local side.

\par The encoding and decoding network of our model is implemented by a GAN (Generative Adversarial Network), which consists of a generator and a discriminator. 
The generator continuously learns the real image distribution and generates a more realistic image to fool the discriminator. 
At the same time, the discriminator needs to discriminate the authenticity of the received images. 
Through the constant confrontation between the generator and the discriminator, they form a min-max game, both sides continuously optimize themselves during the training process until they reach an equilibrium.
Compared with other methods, GAN can achieve compression for full-resolution images and compression for images with extreme code ratios, which has wide applicability. 
Also, the reconstructed images have sharper textures and get better picture results. In our model, the decoder is used as the generator and is trained together with the encoder. 
The customization of the model is very flexible. Besides, it can set the compression ratio by adjusting the feature map size and the number of channels before and after compression. 
When working, the encoder will be reserved locally, and RGB images will be extracted as feature maps for compression and upload. In the cloud, the images will be reconstructed by the decoder.

\par The main contribution of this paper is to propose a safe and efficient multi-object grasping detection scheme for robotic arms. This scheme has three advantages:
\par (1) High fidelity: We have achieved good results on DIV2K, flickr30k, Cornell, and OCID datasets. The compression ratio can be achieved, and the structural loss of the reconstructed image after the transmission is less than 7$\%$, and there is almost no difference in the result of grasp detection before and after compression.
\par (2) Strong security: Transmitting the compressed tensor to the server instead of the original image. This method avoids the leakage of production information or privacy. Compared with traditional image compression algorithms such as JPEG and JP2000, the uploaded data is difficult to be decrypted and is highly reliable. Theoretically, without the corresponding decoder parameters, there is no way to reconstruct the picture even if the transmission information is intercepted.
\par (3) High execution efficiency: First, the local side of the operation is offloaded in the cloud, and the limited local arithmetic power is complemented by the arithmetic power provided by the cloud. Second, the compressed information occupies less bandwidth and is transmitted faster. Third, the lightweight neural network fits the actual application scenario.

\section{Related Work}
The method of achieving automatic grasping of the robotic arm has been improving over the course of long-term research. The traditional methods of perception-based grasping, reconstructing 3D models of objects, and analyzing the geometric features and forces of models, it has gradually expanded to the use of deep learning network models for image object detection and pose estimation\cite{bicchi2000robotic}.
\par The work uses the CNN (FAST R-CNN VGG16) network model to complete the pose estimation after image detection. This work proves the practicality of the object in the case of obscuration through experiments\cite{zhihong2017vision}. Another work proposes a multimodal model method for image detection using ResNet for RGB, which has better performance than VGG16\cite{kumra2017robotic}.
\par Others use deep learning networks to calibrate and control the behaviour of robotic arms.
\par Leoni’s work is based on the RNN network model, through the sensor data to learn and train the robot's grasping behaviour, thus making sure the system can achieve the goals \cite{Leoni1998ImplementingRG}.
\par Several works use RL technology to optimize and train a robot’s gripping ability. After a lot of training, these methods have achieved good experimental results in limited scenes. However, in more complex and practical scenarios, the scalability of RL is still unknown \cite{quillen2018deep}.
\par It is worth noting that the work of Chu et al.\cite{2018Real} on multi-object grasping detection has achieved good results in recent years. Our work is based on the model they proposed.
\par Due to the demand for computing power in deep learning, the use of cloud edge fog computing is also more applied in robot-related fields. For example, in the work of Sarker et al., \cite{sarker2019offloading} the use of offload cloud computing work reduces the energy consumption and hardware requirements of the robot. This treatment reduces a lot of pressure on the hardware part of the robot and the robotic arm.
\par Kumar et al.\cite{tanwani2020rilaas} builds a cloud computing framework. Through this framework, any robot can call the infinite computing power of the cloud to calculate. Deng et al.\cite{deng2016optimal} proposed a set of invocation algorithms for fog computation. This method can allocate resources more reasonably and efficiently in a limited computing power environment that is closer to the actual situation.
\par The processing of cloud edge fog often relies on the stability of the connection and relatively high bandwidth. And in practical application scenarios, the compression of images is an essential part.
\par Some traditional image compression algorithms can achieve certain results in conventional scenes. Dhawan’s summary had already analyzed the advantages and disadvantages of methods such as JPEG. However, this method does not present a good direction for further improvement \cite{dhawan2011review}.
\par Compared with traditional algorithms, the direction of image compression using deep learning has yielded many results.
\par Johannes et al. use the CNN network as a decoder to deal with image compression problems and obtain good theoretical data. This method is processed by the convolutional neural network, which reduces the amount of both computation and image compression data \cite{balle2018variational}. But in the case of practical applications, end-to-end joint optimization is often difficult to complete high-effect compression and high-quality reconstruction of the image at the same time.
\par In addition to this, the limited sensory field of the convolutional kernel makes the training often fail to achieve the expectation. This is because the achievement of full-resolution compression tends to increase the difficulty of training network structures.
\par Toderici et al. use the LSTM network model and the CNN+RNN network model for image compression\cite{Toderici2016VariableRI} \cite{Toderici2017FullRI}. And the network model built by using the LSTM network framework is more robust for different pictures. However, experiments have shown that the training of the model is quite complex. Besides, the image correlation relationship cannot be well grasped, and it can only be limited to small-size pictures.
\par On the other hand, studied the application of VAE networks in image compression. By increasing the mass ratio factor of VAE, linear proportion and other methods achieved a fairly good compression effect \cite{Jiang2021OnlineMA} \cite{Chen2020VariableBI}. However, since VAE networks learn the general and original picture by calculating the mean squared error,  the resulting image is more likely to occur the edge blur problem.
\par Rippel et al. was the first to propose the application of GAN networks to image compression \cite{Rippel2017RealTimeAI}. The decoded data is processed and generated by using a GAN network, and it is opposed to the discriminator supported by the real data. The model can not only complete the compression of full-resolution images but also achieve image compression at a limited bit rate. This results in a reconstructed image with a clear texture for better visual sensory effects.
\par A large number of applications of cloud, edge, and fog computing systems has spawned a very urgent information security problem. In \cite{Joshi2021AnalyticalRO}, for example, the authors analyze the data security issues posed by cloud computing. 
In addition, the review of Randeep and Jagroo \cite{Kaur2015CloudCS} points to the security issues that cloud computing can bring. They summarize techniques for overcoming data privacy issues and define pixel key patterns and image steganography techniques for overcoming data security issues.
\par Some work\cite{Li2020AchievingSA}\cite{Shini2012CloudBM} \cite{Yao2014CloudBasedHI} discussed the security of medical information in cloud storage and data sharing environments and gave some feasible solutions. Overall, these studies highlight the security of information (communications) under the cloud computing system.
\par To sum up, most of the existing robot arm's grasp detection work is highly dependent on their edge computing ability, and the safety problems in the process of grasp detection are not considered enough.

\section{Methodology}

\par
\begin{figure*}
\begin{center}
\includegraphics[height=4.5cm]{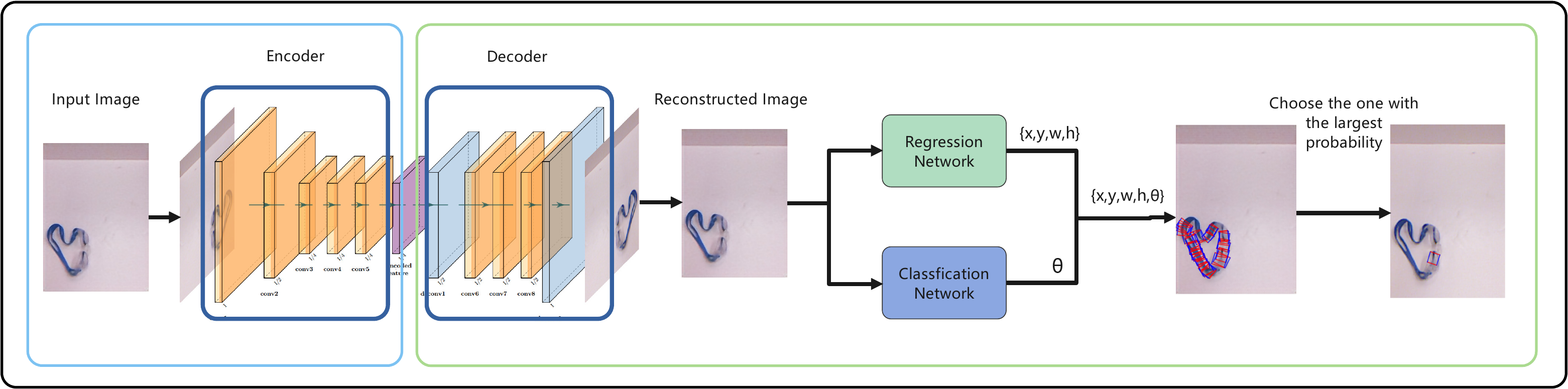}
\end{center}
\caption{The figure shows the general technical flow chart of our approach. The input image is collected at the edge, compressed by the encoder, and then uploaded to the cloud. The image will be reconstructed by the decoder in the cloud. Then grasp parameters are obtained through classification and regression networks to get the bounding box. According to the probability, the most likely bounding box is selected as the final result of grasp detection. The blue box and green box in the figure represent the edge end and cloud end, respectively.}
\label{fig2}
\end{figure*}

 The RGB image is grasped by the local camera and sent out by the edge side after encoding and compression. The cloud side receives the data and then the decoder reconstructs the image for grasp detection. The parameters of the encoder and decoder are obtained using generative adversarial networks for training. Two tasks are completed in the grasp detection phase: grasp proposals and grasp configuration, the former determines the location of the object and the latter configures the grasp angle. The system flowchart is shown in Fig \ref{fig2}. and comprises a number of components, which we will be introduced below. 

\subsection{Image compression part}
\par In this section, we will focus on feature extraction, network architecture design, and customized loss function.

\subsubsection{Feature Extraction and Compression}
\par Our model uses global generative compression for image compression. Before encoding and decoding, the input image is first passed through two layers of convolution to achieve feature extraction and image compression. We found that by adjusting the number of feature channels and feature map size output here, we not only balance the processing speed and image compression quality, but also easily change its compression ratio.
\par We preprocess the image so that the input image is an RGB image with a height of 210 and a width of 150. The encoded image obtained by the encoder is a feature map of 52x37 of 2,4,8,16 channels, the corresponding compression ratio is 32.58$\%$, 16.29$\%$, 8.14$\%$ and 4.07$\%$, respectively. The calculation of the compression ratio is given by Equation \eqref{eq1}. It represents the ratio of the parametric quantities of  the output tensor  $R^{C_{c}\times H_{c}\times W_{c}} $  to the input image  $R^{C_{i}\times H_{i}\times W_{i}} $   . The reconstructed images are similar to the original images, whose structural similarity index is greater than 0.93. 

\begin{equation}\label{eq1}
\boldmath
Compression\ ratio= \frac{C_{c} \times H_{c} \times W_{c} } {C_{i}  \times H_{i} \times W_{i} }\times 100\%
\end{equation}

\par
\begin{figure}[!htbp]
\centerline{\includegraphics[height=6cm]{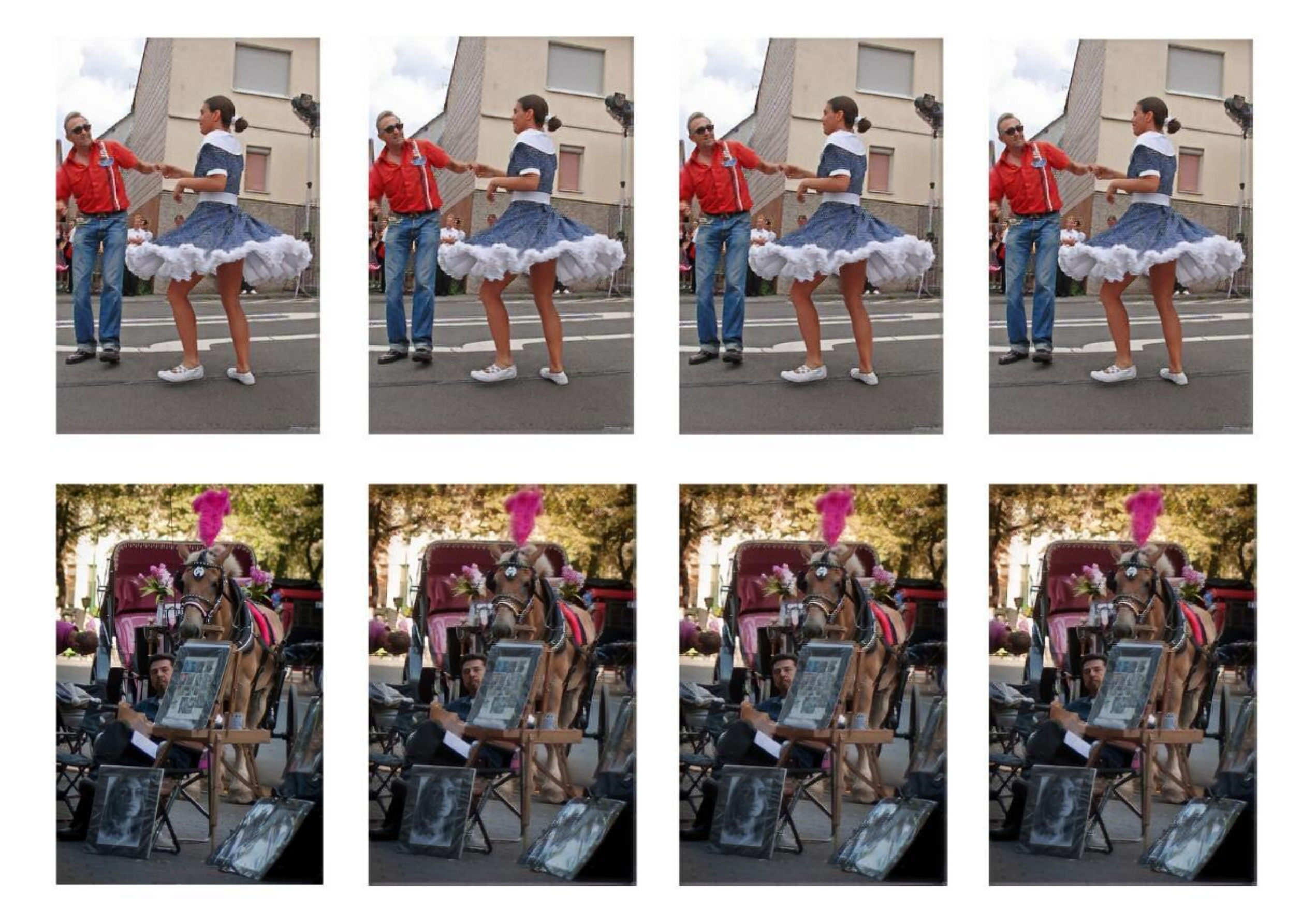}}
\caption{Results of reconstructed images. The left column is the original image, and the right three columns are the reconstructed images with 32.58\%, 16.29\%, and 8.14\% compression respectively}
\label{fig3}
\end{figure}

 The number of parameters for different compression ratios is shown in Table 1, and the detailed results under different compression ratios will be given in the experimental section. In Figure \ref{fig3}, we show the reconstructed image results under different compression ratios.

\renewcommand\arraystretch{1.5}  
\begin{table}[!htbp]
	\centering
        \setlength{\tabcolsep}{4mm}
	\caption{Number of parameters before and after image compression}
	\begin{tabular}{ccc}
		\toprule  
		\tabincell{c}{ Input image }&\tabincell{c}{Compressed tensor}&\tabincell{c}{Compression ratio}\\
		\midrule  
		\tabincell{c}{94500}&30784&32.58$\%$ \\ 
		\midrule  
		\tabincell{c}{94500}&15392&16.29$\%$ \\
		\midrule
		\tabincell{c}{94500}&7296&8.14$\%$\\
            \midrule
		\tabincell{c}{94500}&3648&4.07$\%$\\
		\bottomrule  
	\end{tabular}
\end{table}

\subsubsection{Network Architectures}
\par In order to make the network structure as simple as possible, here we have built a lightweight Generator advertising network that is similar to DCGAN \cite{Radford2016UnsupervisedRL}. The network consists of a Generator and a Discriminator. It uses a decoder as a generator and trains the encoder and decoder by using the same loss function in training. During training, the goal of the generator is to try to generate real images to deceive the discriminator. And the goal of the discriminator is to try to separate the images generated by the generator from the real images and then paste the 0 and 1 labels respectively.
\par After the encoding stage, we only upload the tensor generated by the encoding network to the cloud without anything else. On the cloud, we use the decoder to restore the tensor to a reconstructed image. In the encoder (compressor network), we used three consecutive layers of simple residual layers (ResNet\cite{He2016DeepRL}) for encoding. Correspondingly, in the decoder (decompressor network), the two upsample and three layers of residual layers are crossed, and eventually received a reconstructed image. We implemented upsample with transposed convolution and restored the dimensions of the output picture. In the encoding and decoding network, we use LeakyReLU as the activation function and use Tanh in the last layer. In the convolution block during the encoding and decoding phases, we keep the size of the feature map constant by setting the stride and padding, which reduces the loss of information. For the discriminator, we built a simple model based on a combination of convolution and dropout layers.

\subsubsection{Loss Function}
\par Generally in GAN, we tend to use L1 loss (MAE) and L2 loss (MSE) to train discriminators for binary classification problems. However, it cannot be ignored that the simple use of L1 loss for judgment often fails to accurately reflect the level of detail of the image compression and restoration. Structural loss is also a structural loss consideration in image compression tasks. To consider both, we divide the loss function into two parts, namely, the adversarial loss and the structural loss weighting add up to the final loss function.
\par There are many types of loss functions based on deep learning image algorithms, such as L1 loss and L2 loss. However, for image compression and restoration work, these two loss functions are not easy to recover for the detailed structure of the image and are not enough to intuitively express people's cognitive feelings. In addition, there is also PSNR (Peak Signal-to-Noise Ratio) as a common evaluation criterion, but it has a common problem with L1 and L2: their principle is based on pixel-by-pixel comparison differences, without considering human visual perception, so the PSNR index is high, not necessarily representing image quality.
\par So here we use MSSIM\cite{wang2004image} as a structural loss, which is based on SSIM. SSIM is a commonly used image quality evaluation index, which is based on the assumption that the human eye will extract the structural similarity variables when viewing the image. Its final loss value is obtained by comprehensively considering the brightness, contrast, and structural similarity variables. For images x and y, their SSIM is calculated as follows:
\begin{equation}\label{eq2}
\boldmath
l(x,y)=\frac{2\mu_{x}\mu_{y}+C_{1}}{\mu_{x}^{2}+\mu _{y}^{2}+C_{1}}
\end{equation}
\begin{equation}\label{eq3}
\boldmath
c(x,y)=\frac{2\sigma_{x}\sigma_{y}+C_{2}}{\sigma_{x}^{2}+\sigma_{y}^{2}+C_{2}}
\end{equation}
\begin{equation}\label{eq4}
\boldmath
s(x,y)=\frac{\sigma_{xy}+C_{3}}{\sigma_{x}\sigma_{y}+C_{3}}
\end{equation}

\par In Equation \eqref{eq2} \eqref{eq3} \eqref{eq4},  $l(x, y)$  is used to estimate luminance by mean, $c(x,y)$ is used to estimate contrast with variance, and $s(x,y)$  is used to estimate structural similarity with covariance. The SSIM definition is shown in Equation \eqref{eq5}, where $\alpha$, $\beta$, and $\gamma$ are used to adjust the weights of each portion. By default, we set all three of them to 1, and then we can get the Equation\eqref{eq6}

\begin{equation}\label{eq5}
\boldmath
SSIM(x,y)=[l(x,y)]^{\alpha }\cdot[c(x,y)]^{\beta }\cdot[s(x,y)]^{\gamma}
\end{equation}

\begin{equation}\label{eq6}
\boldmath
SSIM(x,y) = \frac{(2\mu_{x}\mu_{y}+C_{1})(2\sigma_{xy}+C_{2})}{(\mu_{x}^{2}+\mu_{y}^{2}+C_{1})(\sigma_{x}^{2}+\sigma_{y}^{2}+C_{2})} 
\end{equation}

\par MSSIM takes the reference image and the distortion image as input and divides the image into N blocks by sliding window. Then it will weight the mean, variance and covariance of each window, and the weight $w_{ij}$ meets the $\sum_{i=0}^{n}\sum_{j=0}^{n} w_{ij} = 1$. 
We usually use the Gaussian kernel to calculate the structural similarity SSIM of the corresponding block and use the average value as a final structural similarity measure of the two images. Let’s suppose the original image is scale 1, and the highest scale is scale M obtained by the M-1 iteration. For the $j^{th}$ scale, only the contrast $c(x,y)$ and the structural similarity $s(x,y)$ will be calculated. Brightness similarity $l(x,y)$ is calculated only in Scale M. The final result is to link the results of the various scales:

\begin{equation}\label{eq7}
\begin{split}
\boldmath
{MSSIM}(x, y)= \left[l_{M}(x, y)\right]^{\alpha_{M}} \\ \cdot  \prod_{j=1}^{M}\left[c_{j}(x, y)\right]^{\beta_{j}} \cdot\left[s_{j}(x, y)\right]^{\gamma_{j}}
\end{split}
\end{equation}

\par Hang et al.\cite{Zhao2017LossFF} demonstrates the quality of these loss functions through three experiments. It shows that MSSIM is more appropriate in comparison. In order to make the output image gain higher quality and easier training, here we use the loss function combined with MSSIM and L1 loss.

\subsection{Grasp detection part}
\par The entire grasp detection task is divided into two tasks, grasp proposals, and grasp configuration. The former determines the location of the object, and the latter configures the angle of the grasp.

\subsubsection{Grasp Proposals}
\par Grasp proposals are implemented by using the two-stage detection algorithm and consist of two branches: regression and classification. The model chose the ResNet-50 network as the backbone of model. First, the location of the bounding box is determined by regression, which generates the region proposals, avoiding the time-consuming sliding window method and directly predicting the region proposals on the entire image.
\par These region proposals will make feature extraction through RPN (Region Proposal Network)\cite{ren2015faster}, and the region frame proposals classification is completed when the region frame proposals extraction is performed. The classification process classifies region features into background and object.
\par When the RPN network generates a region proposal, the position of the object is preliminarily predicted. During this time, the two links of regional classification and location refinement are completed. As soon as it obtains the region proposals, the ROI pooling layer will accurately refine and regress the position of the region proposals.
\par After the region target corresponds to the features on the feature map, the characteristics of the region proposals will be further represented through a fully connected layer. Later, the category of the region target and the refinement of the region target position will be completed by classification and regression, so the real category of the object will be obtained. While the regression will get the specific coordinate position of the current target, which is represented as a rectangular box represented by four parameters.

\subsubsection{Grasp Configuration}
\par The determination of the grasp configuration is achieved through classification. Grasp orientation coordinate $\theta$ divides the direction of the grasp into 20 classes and chooses the class directly with the highest confidence level to grasp.
\par There is a non-grasp direction class in classes. If the confidence level of the output is lower than that of the non-grasp direction class, this grasp recommendation is considered to be ungraspable in that direction. Setting non-grasp classes instead of setting specific thresholds will be a better way to handle multi-object and multi-grasp component tasks.
\par
\begin{figure}[!htbp]
\begin{center}
\includegraphics[height=4cm]{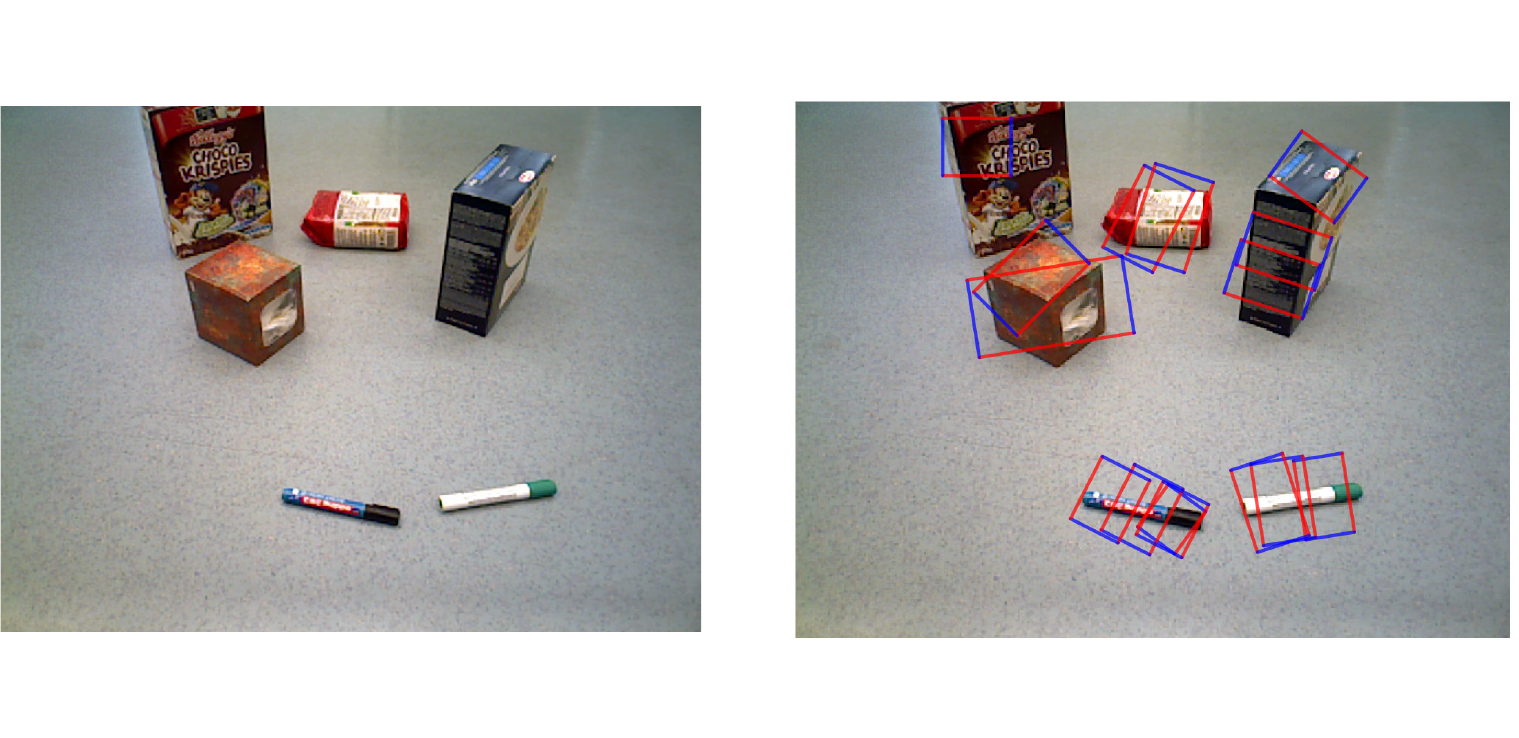}
\end{center}
\caption{Results of grasp detection output. The red line represents the open length of a two-fingered gripper, while the blue line represents the parallel plates of the gripper.}
\label{fig4}
\end{figure}

 The final output is shown in Fig \ref{fig4}. In the figure’s output bounding box, the red line represents the open length of a two-fingered gripper, while the blue line represents the parallel plates of the gripper.

\par In this scheme, the loss function is designed to be two parts: the grass proposal loss $L_{gpn}$ and the grasp configuration loss $L_{gcr_{-}reg}$. As shown in Equation \eqref{eq7}, $L_{gp_{-}cls}$ is the cross-entropy loss of the grasp direction classification, and the $L_{gp_reg}$ and weight $\lambda$ are the L1 regression loss of the grasp recommendation. In the case of no grasping, $p_{i}^{\ast } = 0$. Correspondingly, the $p_{i}^{\ast } = 1$ when it can be grasped. The parameter, $t_{i}^{\ast }$ , and $p_{i}^{\ast }$ are corresponding to the ground truth.

\begin{equation}\label{eq8}
\boldmath
\begin{split}
L_{g p n}\left(\left\{\left(p_{i}, t_{i}\right)_{i=1}^{\mathbf{I}}\right\}\right)= 
\sum_{i}  L_{g p_{-} c l s}\left(p_{i}, p_{i}^{\ast}\right) \\
+ \lambda \sum_{i} p_{i}^{\ast} L_{g p_{-} r e g}\left(t_{i}, t_{i}^{\ast}\right) .
\end{split}
\end{equation}

\par Equation \eqref{eq8} defines the loss function that executes the fetch configuration prediction. In this equation, $L_{gp_{-}cls}$  is the cross-entropy loss of the grasp orientation classification, and $\rho_{l}$ is the confidence level of each classification. $L_{gcr_{-}reg}$ is the regression loss of the bounding box, and the $\beta_{c}$ records the corresponding prediction of the grasp bounding box. $\beta_{c}^{\ast}$ is the correct bounding box. $\lambda_{2}$ is the relative weight.

\begin{equation}\label{eq9}
\boldmath
\begin{split}
L_{g p n}\left(\left\{\left(\rho_{l}, \beta_{l}\right)_{i=1}^{\mathbf{C}}\right\}\right)= 
\sum_{i}  L_{gcr_{-} c l s}\left(p_{l}\right) \\ 
+ \lambda_{2} \sum_{c}1_{c\neq 0}(c) L_{gcr_{-} r e g}\left(\beta_{i},\beta_{i}^{\ast}\right) .
\end{split}
\end{equation}
\\ \hspace*{\fill} \\

\par The total loss consists of the addition of $L_{gpn}$ and $L_{gcr_{-}reg}$, as shown in Equation \eqref{eq9}:

\begin{equation}\label{eq10}
\boldmath
L_{\text {total }}=L_{g p n}+L_{g c r}.
\end{equation}
\\ \hspace*{\fill} \\

\section{Experimental}
\subsection{Experimental Environment}
\par The training environment of the model is an Intel (R) Xeon (R) platinum 8255c, 47GB memory, 12 cores  computer equipped with 24g video memory  GeForce RTX™ 3090 graphics card. The computer system environment is Ubuntu 20.04 operating system. Later, the test experiment was conducted on another GeForce RTX ™ 2080 Ti graphics card.

\subsection{Dataset and Data Preprocessing}
\par We used the Flickr30k \cite{Jia2015GuidingLT} dataset alone for training the image compression reconstruction, and then validated on all four datasets, Flickr30k, Div2k\cite{Agustsson_2017_CVPR_Workshops}, Cornell\cite{ainetter2021end}, and OCID\cite{suchi2019easylabel}. The image reconstruction achieved good results on both PSNR and SSIM values. The grasping training and validation were then performed using the OCID dataset with 92\% accuracy in general.

\par Flickr30k: Flickr30k is the first image description dataset that contains 158,915 descriptions and 31,783 images. This dataset is based on the previous Flickr8k dataset and focuses on describing everyday human activities. Of these, 25,426 images were used for training, and 6,357 images each for validation and testing.

\par Div2k: The DIV2K dataset is a commonly used dataset for super-resolution image reconstruction. The dataset contains 1000 2K resolution images, including 800 training images, 100 validation images, and 100 test images. And the low-resolution images with 2, 3, 4, and 8 reduction factors are provided.

\par Cornell: The Cornell grasping dataset is a required dataset for robotic autonomous grasping tasks. The dataset contains 885 RGB-D images of 640 × 480px size with 240 graspable objects. The correct grasping candidate is given by a manually annotated rectangular box. Each object corresponds to multiple images with different orientations or poses, and each image is labelled with multiple ground truth grasps, corresponding to the many possible ways of grasping the object.

\par OCID: We use the OCID\_grasp dataset part, which is composed of 1763 selected RGB-D images, of which there are more than 75,000 hand-annotated grasp candidates.

\subsection{Training schedule}
\par We train the whole network for 10 epochs on a single GeForce RTX™ 3090. The initial learning ratio is set to 0.0002. The batch size is set to 30, and the log is output every 50 batches The input image is first cropped to 210 × 160 sizes.

\subsection{Evaluation Metric}
\subsubsection{Compressed image quality metrics}
\par PSNR (Peak Signal to Noise Ratio) PSNR is defined as Equation (\ref{eq11}), $MAX_I^{2}$ which is the maximum possible pixel value of the image, and MSE is the mean square error of each pixel point of the two images. the minimum value of PSNR is 0, and the larger the PSNR, the smaller the difference between the two images. We test 100 images and finally take the average as the final value.
\begin{eqnarray} \label{eq11}
P S N R & = & 10 \cdot \log _{10}\left(\frac{M A X_{I}^{2}}{M S E}\right)
\end{eqnarray}
\par SSIM (Structure Similarity Index Measure) Equation (\eqref{eq6}) is the definition of SSIM. SSIM is based on the assumption that the human eye extracts structured information from an image and integrations the differences between two images in terms of luminance, contrast, and structure. $SSIM\le 1$, the larger the SSIM, the more similar the two images are. We test 100 images and finally take the average as the final value.
\subsubsection{Grasping accuracy metrics}
\par The accuracy of the grasping parameters is evaluated by comparing the closeness of the grasp candidate to ground truth.

\par A grasp candidate is considered as a successful grasp detection after satisfying the following two metrics,
\par (1) The difference between the angle of predicted grasp $g_{p} $ and ground truth $g_{t} $ does not exceed 30°.
\par (2) Intersection over Union (IoU) of $g_{p} $ and $g_{t} $ is greater than 25$\%$, which means 
\begin{eqnarray} \label{eq12}
I o U & = & \frac{\left|g_{p} \cap g_{t}\right|}{\left|g_{p} \cup g_{t}\right|}>0.25
\end{eqnarray}

\subsection{Comparative experiment}
\subsubsection{Image Compression Quality Experiment}
\par We conducted compression encoding experiments on the pictures of Flickr30k, Cornell, and DIV2K datasets respectively. The encoding tensor sizes obtained under different datasets and different compression ratios are shown in Table 2. The data in Table 2 show that the magnitude of the compression tensor is proportional to the compression ratio. And satisfies the previously derived formula to present a linear relationship.

\renewcommand\arraystretch{1.5}  
\begin{table}[!htbp]
	\centering
        \setlength{\tabcolsep}{4mm}
	\caption{Output tensor size of the encoder with different compression ratios}
	\begin{tabular}{cccc}
		\toprule  
		\tabincell{c}{ Compression Ratio }&\tabincell{c}{  Flickr30k }&\tabincell{c}{ Cornell }&\tabincell{c}{DIV2K}\\
		\midrule  
		\tabincell{c}{2.03\%}&15.5kb&74.8kb&74.8kb\\ 
		\midrule  
		\tabincell{c}{4.07\%}&30.3kb&148kb&148kb \\
		\midrule
		\tabincell{c}{8.14\%}&60kb&297kb&297kb\\
		\midrule
        \tabincell{c}{16.29\%}&119kb&593kb&593kb\\
		\midrule
         \tabincell{c}{32.58\%}&237kb&1187kb&1187kb \\
		\bottomrule  
	\end{tabular}
\end{table}

\par We select 200 images from each of the three datasets of Flickr30k, Cornell, and DIV2K, and divide them into 2:3 batches according to the complexity of the images. The reconstructed image is compared with the original image at different compression ratios. The reconstructed image is compared with the original image at different compression ratios. We get their PSNR and SSIM values and average them to get Table 3 and Table 4. The data in tables 3 and 4 show that our model has achieved good results under picture input of different complexity. The average value of PSNR and SSIM reached 35.576 and 0.948 respectively

\renewcommand\arraystretch{1.5}  
\begin{table*}[]
\centering
\setlength{\tabcolsep}{1.3mm}
\caption{PSNR of different dataset}
\label{tab:my-table}
\begin{tabular}{@{}cclccclccc@{}}
\toprule
\multicolumn{1}{c}{Compression} & \multicolumn{3}{c}{DIV2K} & \multicolumn{3}{c}{Flickr30K} & \multicolumn{3}{c}{Cornell} \\ \cline{2-10} 
\multicolumn{1}{c}{Ratio} & High complexity & Low complexity & Avg & High complexity & Low complexity & Avg & High complexity & Low complexity & Avg \\ \midrule
32.58\% & 24.48 & 30.31 & 27.98 & 21.31 & 29.78 & 26.392 & 34.785 & 36.10 & 35.5 \\
16.29\% & 24.885 & 30.26 & 28.108 & 22.29 & 29.56 & 26.652 & 31.41 & 32.01 & 31.768 \\
8.14\% & 24.16 & 29.31 & 27.248 & 21.82 & 28.31 & 25.714 & 29.93 & 30.41 & 30.216 \\
4.07\% & 22.265 & 27.98 & 25.692 & 19.7 & 25.85 & 23.39 & 31.57 & 32.66 & 32.226 \\
2.03\% & 19.515 & 24.44 & 22.472 & 19.52 & 24.44 & 22.472 & 26.875 & 27.91 & 27.496 \\
0.99\% & 18.255 & 22.67 & 20.9 & 17.31 & 22.34 & 20.328 & 29.535 & 30.44 & 30.078 \\
0.50\% & 15.55 & 20.51 & 18.528 & 15.91 & 19.02 & 17.776 & 24.38 & 26.86 & 25.866 \\
0.13\% & 15.175 & 18.92 & 17.422 & 14.90 & 18.45 & 17.03 & 22.49 & 24.26 & 23.552 \\
0.06\% & 16.225 & 21.21 & 19.218 & 15.46 & 18.49 & 17.276 & 19.475 & 19.11 & 19.254 \\
0.04\% & 13.09 & 16.41 & 15.084 & 14.26 & 17.91 & 16.446 & 11.915 & 11.43 & 11.626 \\
0.03\% & 10.875 & 14.99 & 13.344 & 12.24 & 13.59 & 13.05 & 11.915 & 11.43 & 11.626 \\
\bottomrule  
\end{tabular}
\end{table*}

\renewcommand\arraystretch{1.5}  
\begin{table*}[]
\centering
\setlength{\tabcolsep}{1.5mm}
\caption{SSIM of different dataset}
\label{tab:my-table}
\begin{tabular}{@{}cccccclccc@{}}
\toprule
\multicolumn{1}{c}{Compression} & \multicolumn{3}{c}{DIV2K} & \multicolumn{3}{c}{Flickr30K} & \multicolumn{3}{c}{Cornell} \\ \cline{2-10} 
\multicolumn{1}{c}{Ratio} & High complexity & Low complexity & Avg & High complexity & Low complexity & Avg & High complexity & Low complexity & Avg \\ \midrule
32.58\% & 0.83 & 0.86 & 0.86 & 0.75 & 0.85 & 0.81 & 0.945 & 0.95 & 0.948 \\
16.29\% & 0.835 & 0.89 & 0.868 & 0.78 & 0.86 & 0.828 & 0.945 & 0.95 & 0.948 \\
8.14\% & 0.805 & 0.87 & 0.846 & 0.75 & 0.82 & 0.792 & 0.945 & 0.95 & 0.948 \\
4.07\% & 0.66 & 0.81 & 0.752 & 0.64 & 0.87 & 0.778 & 0.935 & 0.94 & 0.938 \\
2.03\% & 0.48 & 0.7 & 0.612 & 0.51 & 0.78 & 0.672 & 0.905 & 0.92 & 0.914 \\
0.99\% & 0.545 & 0.61 & 0.586 & 0.47 & 0.59 & 0.542 & 0.885 & 0.89 & 0.888 \\
0.50\% & 0.365 & 0.46 & 0.422 & 0.49 & 0.6 & 0.556 & 0.845 & 0.87 & 0.858 \\
0.13\% & 0.33 & 0.43 & 0.39 & 0.35 & 0.48 & 0.428 & 0.825 & 0.83 & 0.83 \\
0.06\% & 0.405 & 0.49 & 0.458 & 0.37 & 0.53 & 0.466 & 0.77 & 0.80 & 0.79 \\
0.04\% & 0.27 & 0.34 & 0.312 & 0.305 & 0.478 & 0.408 & 0.555 & 0.56 & 0.556 \\
0.03\% & 0.225 & 0.29 & 0.262 & 0.23 & 0.35 & 0.302 & 0.555 & 0.56 & 0.556 \\ 
\bottomrule  
\end{tabular}
\end{table*}

\par In Tables 3 and 4, we use eleven compression ratios to test the image compression and reconstruction effects under different compression ratios. The results show that when the compression ratio is above 4.07\%, the accuracy will not decrease too much with the decrease of the compression ratio. When the compression ratio is 2.03\% or less, the loss gradually manifests. The results show that our model has a strong feature extraction ability and a large range of customizable compression ratios.

\par With the increase of the compression ratio, the image reconstruction quality increases sub-linearly and finally tends to a higher value. From the comparison of eleven groups of values, it can be seen that by weighing the compression ratio and image quality, from the image reconstruction quality index, 8.14\% and 16.29\% are the best compression ratio settings of the network.  The data in the table shows that the SSIM value of image reconstruction on the three datasets is greater than 0.82 under these two radios. The Cornell dataset in the actual grasping environment has the highest score, with a PSNR of 31.768 and an average SSIM of 0.948, which is sufficient to meet the needs of grasping. However, in the actual process of grasping and detecting, the requirements for images are not the same as those for human eyes. We will conduct further experiments in combination with grasping in the two experiments of the grasp detection accuracy experiment and the network architecture experiment.

\begin{figure}[!h]
\begin{center}
\includegraphics[height=6cm]{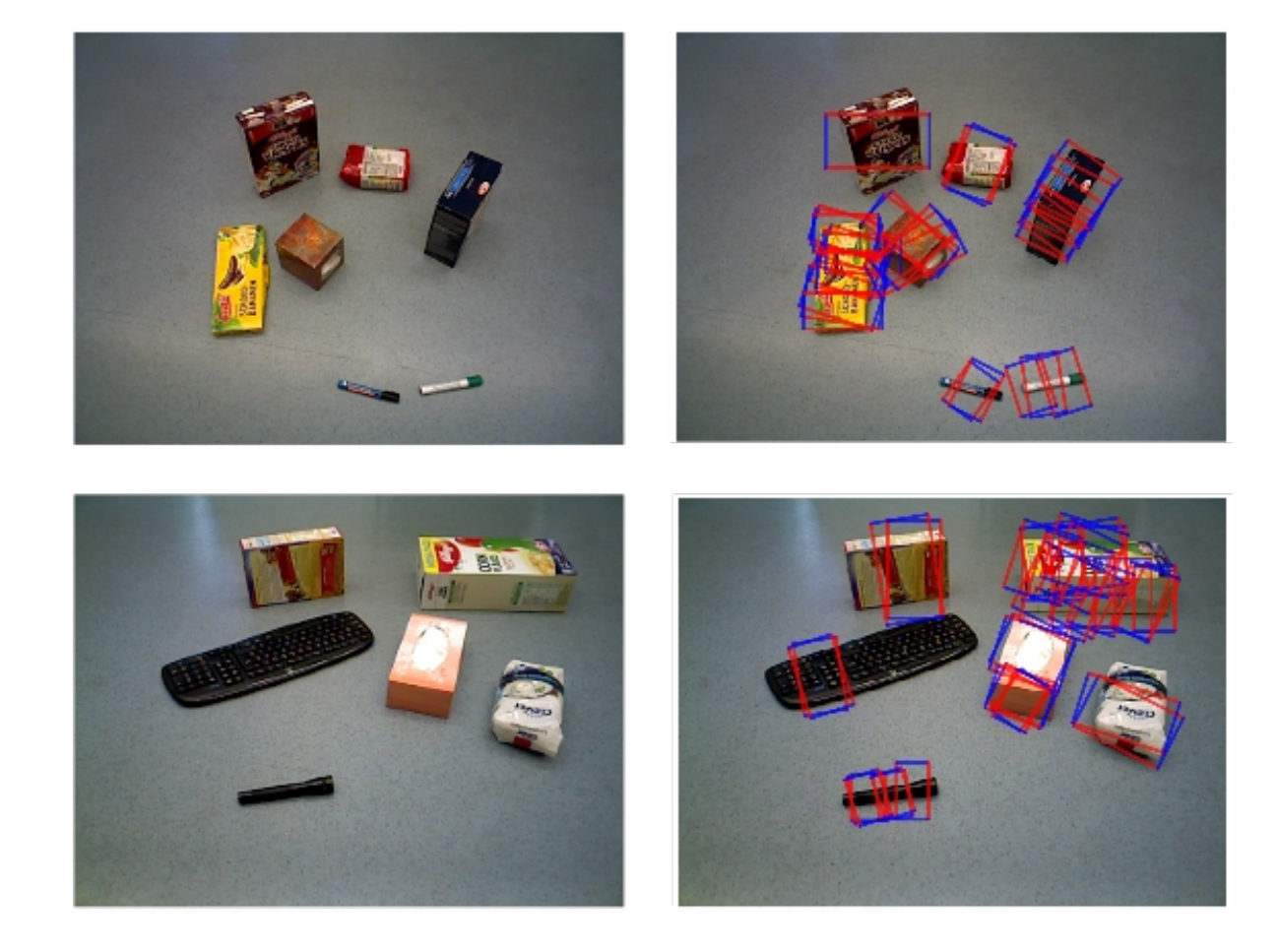}
\end{center}
\caption{Performance of our model on OCID dataset. In the output bounding box, the red line represents the open length of a two-fingered gripper, while the blue line represents the parallel plates of the gripper}
\label{fig5}
\end{figure}

\subsubsection{Grasp Detection Accuracy Experiment}
\par
\begin{figure}
\begin{center}
\includegraphics[height=9.5cm]{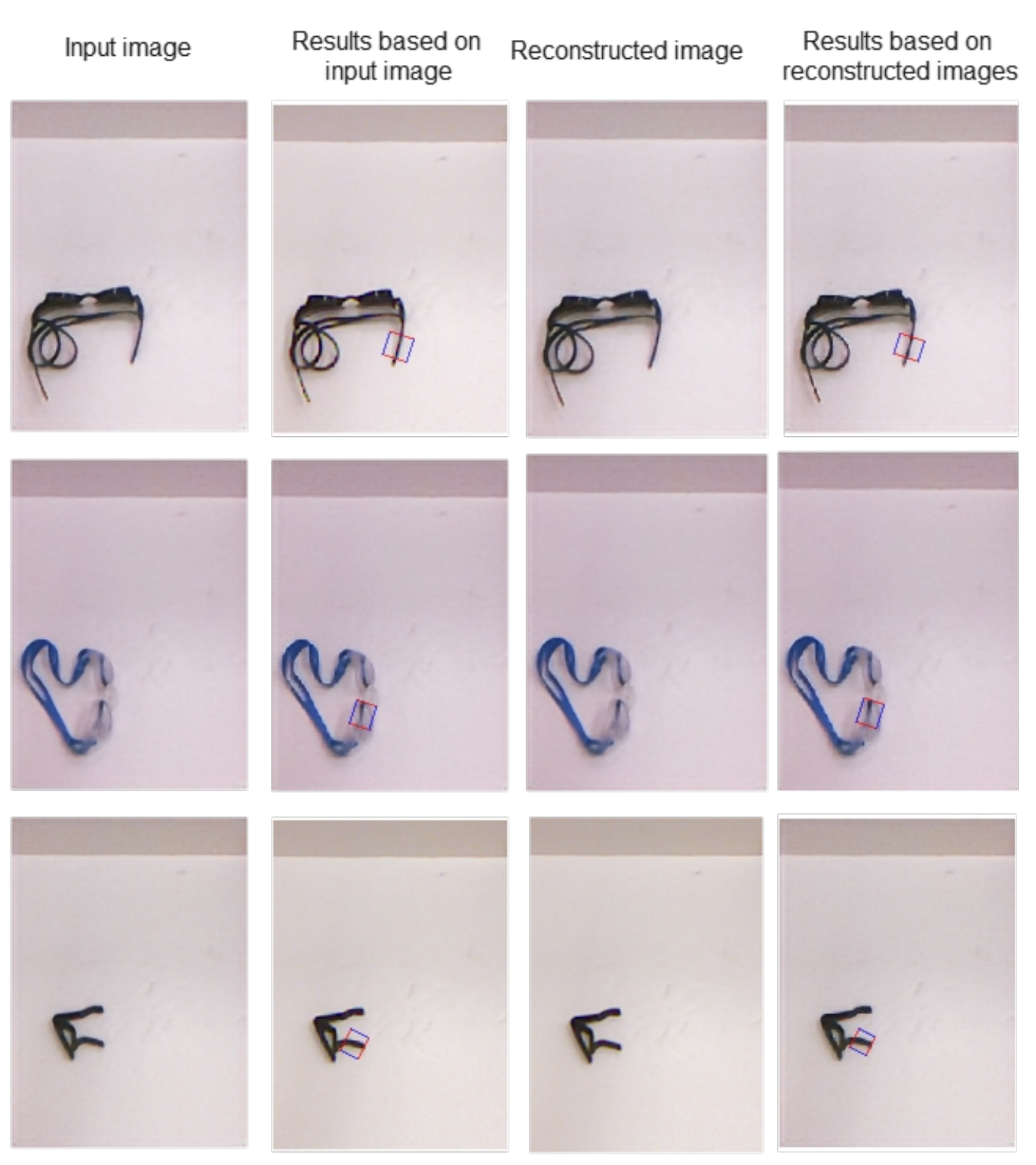}
\end{center}
\caption{The first column is the original input pictures. The second column is the result of grasp detection based on the original image. The third column shows the compressed and reconstructed image. The fourth column is the result of the grasp detection based on the compressed and reconstructed image. It can be seen from the figure that the loss of compression accuracy is little, and grasping accuracy has not been greatly affected.}
\label{fig6}
\end{figure}

 In order to evaluate the effect of encoding and decoding on grasp detection, we compared the results of grasping detection using the original image and the reconstructed image.  The results are shown in Fig \ref{fig6}.

\par We can see from Fig \ref{fig6} that under the compression ratio of 8.14\%,  the accuracy does not decrease too much after being compressed. At the same time, the processing speed of our grasp detection algorithm can reach 13.62fps in the implementation environment.

\begin{figure}
\begin{center}
\includegraphics[height=5cm]{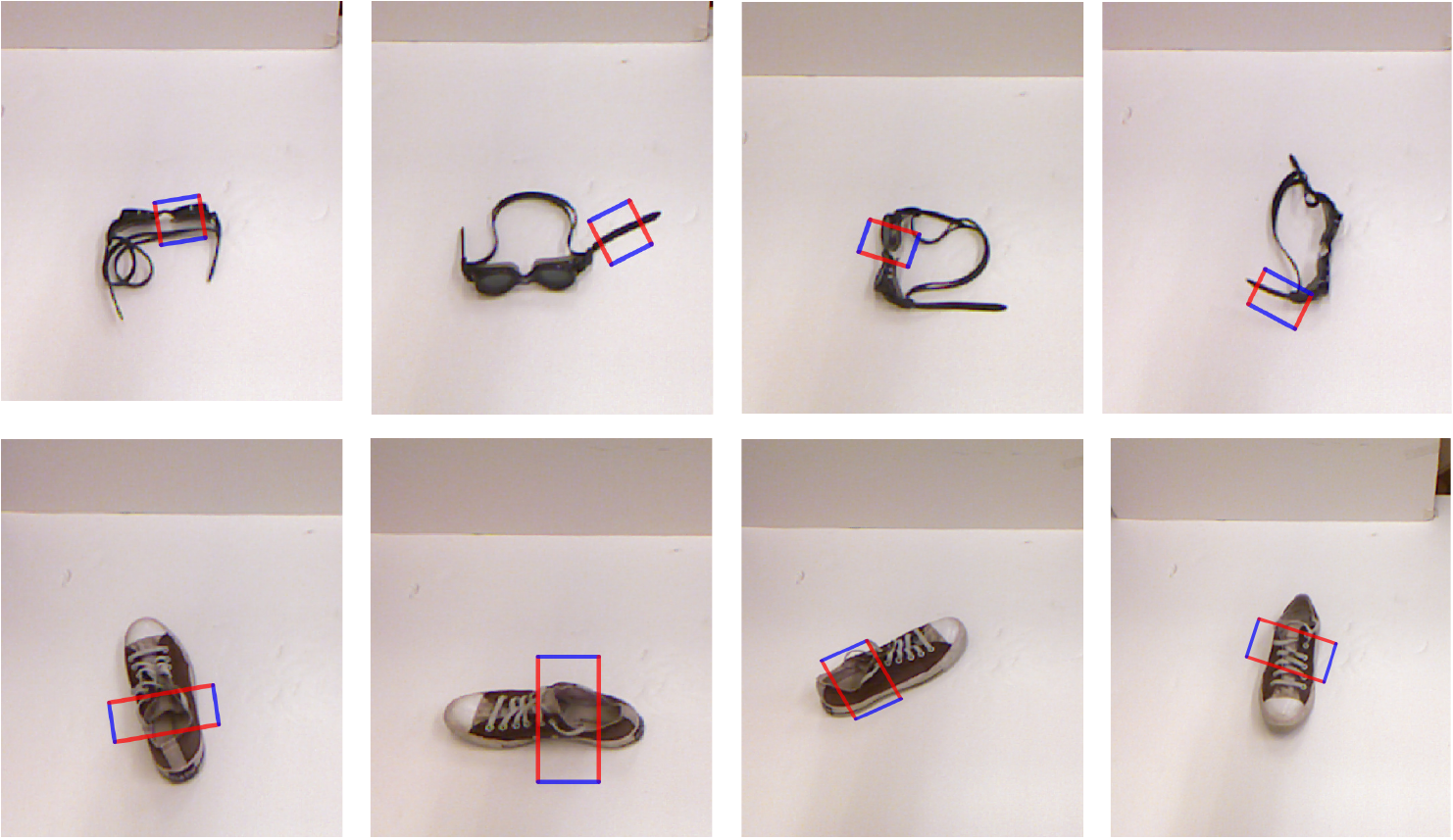}
\end{center}
\caption{Grasp detection experiment based on the same object from multiple angles. The figure shows that our model can accurately mark the bounding box in different directions.}
\label{fig7}
\end{figure}

\par The Cornell dataset provides images and grasp labels from multiple angles of each object. We carry out the grasp detection experiment based on the same object from multiple angles. Fig.\ref{fig7} shows the effect. Our model can accurately mark the bounding box at different angles.

\par We detect the accuracy of multi-object grasp task in the environment of a single object, less than ten objects, and more than ten objects. Count the number of successful grasp detection and calculate the grasp accuracy. The results are shown in Table 4. The results show that when the number of objects is less than 5, our model can basically achieve 100\% error-free detection on the OCID dataset. Fig \ref{fig5} shows the performance of our model on the OCID dataset.

\begin{figure*}
\begin{center}
\includegraphics[height=11cm]{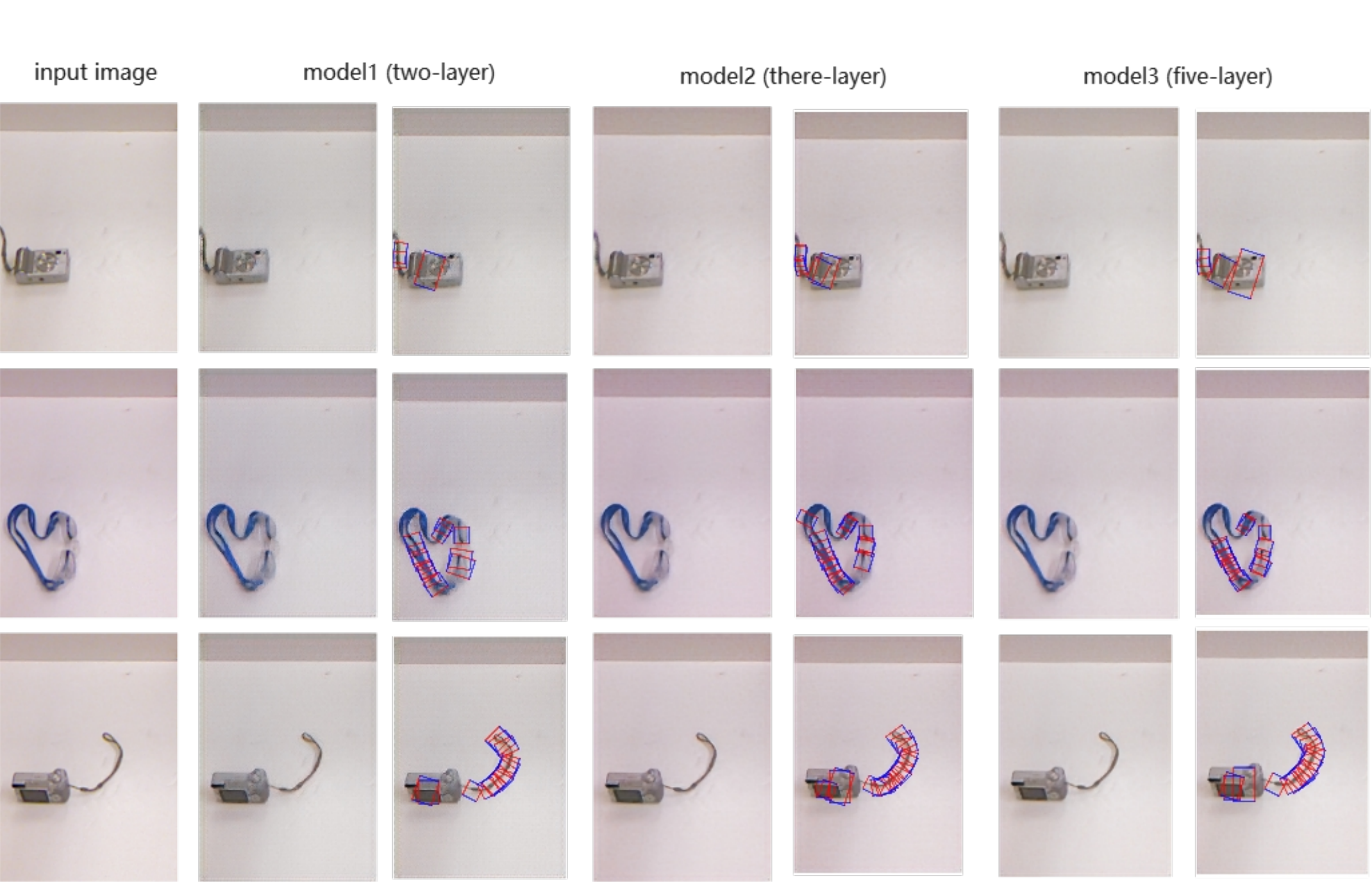}
\end{center}
\caption{Reconstruction and grasp detection result of different models. As the number of convolution layers increases, feature extraction becomes more sufficient, and the quality of reconstructed pictures becomes better.}
\label{fig8}
\end{figure*}

\subsubsection{Network Architectures Experiment}
\par In order to reasonably design the parameters of the neural networks, we carried out parameter optimization experiments from two dimensions of network depth and the number of channels.

\par We designed different models with two, three, and five convolution blocks of network architectures respectively for image reconstruction experiments. The results are shown in Fig.\ref{fig8} and Table 5, to compare the effect of the number of encoder-decoder network layers on the model effect.
By comparing these figures and tables,  it can be concluded that the reconstructed image of the three-layer convolution block model is better than that of the two-layer coding block network. However, for the five-layer network, considering the operation speed and guarantee ratio, we think that three layers are the better network layers.

\par Comparison of reconstructed image quality under different channel numbers is shown in Fig.\ref{fig12}-\ref{fig16} in the appendix.  The image is blurred at a low compression ratio, but it can still be detected and judged. With the increase in compression ratio, the result of grasp detection is close to the original image. The three lines from top to bottom are the input image, reconstructed image, and result with the label. It can be concluded from these five groups of pictures that the compression ratio of 0.13\% and above is similar to the original picture, which can ensure the accuracy of grasping.

\renewcommand\arraystretch{1.5}  
\begin{table}[]
\centering
\caption{Comparison of different models}
        \setlength{\tabcolsep}{7mm}
\begin{tabular}{ccc}
		\toprule  
Number of convolution layers & PSNR & SSIM \\
		\midrule  
2-layer & 28.346 & 0.9 \\
		\midrule  
3-layer & 31.768 & 0.948 \\
		\midrule  
5-layer & 35.438 &0.952\\
		\bottomrule  
\end{tabular}
\end{table}

\begin{figure*}
\begin{center}
\includegraphics[height=19.5cm]{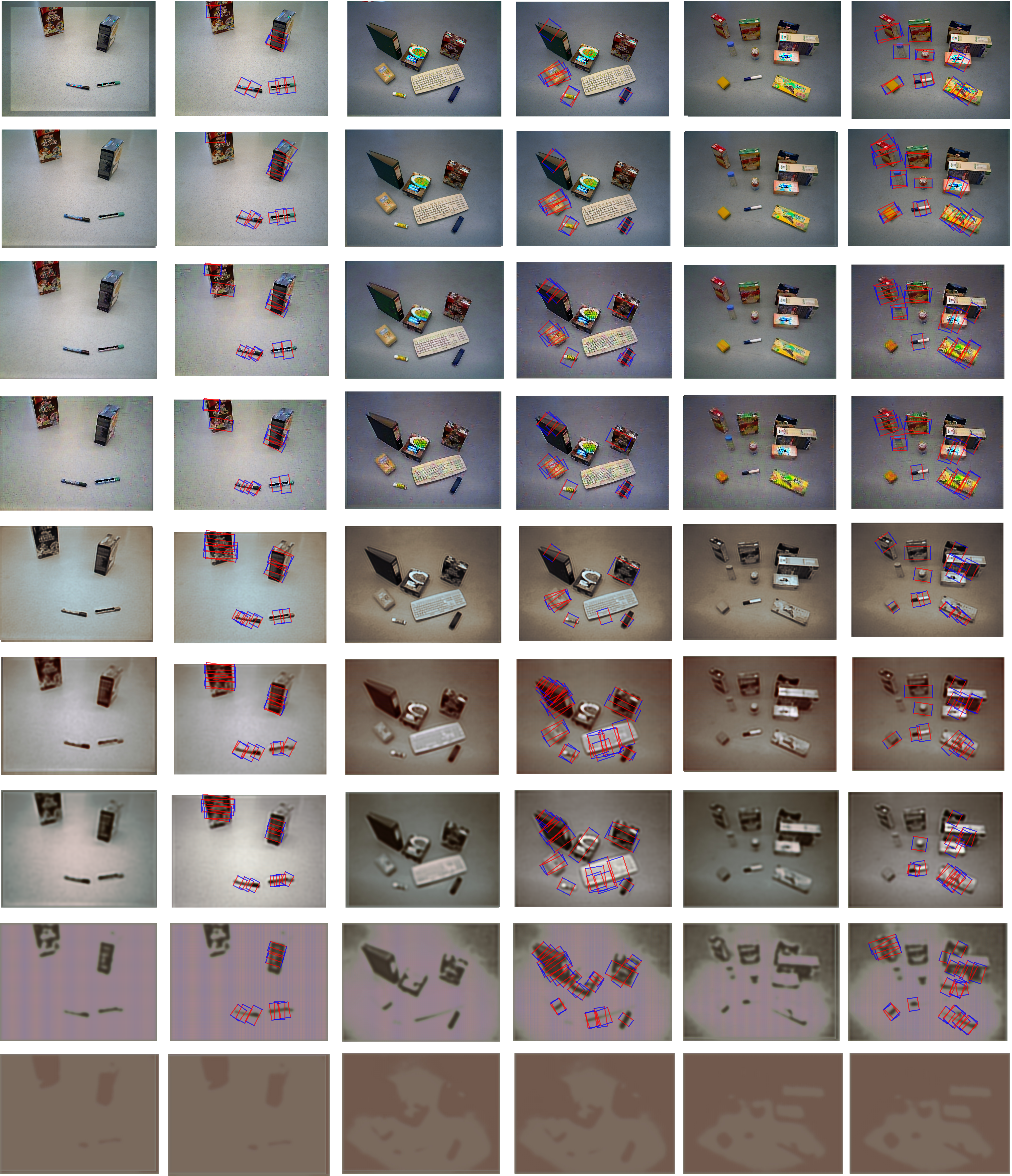}
\end{center}
\caption{The first line is the result of the grasp detection of the original image, and the lines below are the results under the compression ratio of 16.29\%, 4.07\%, 0.99\%, 0.5\%, 0.13\%, 0.05\%, 0.04\%, and 0.03\%. From top to bottom, the loss of pictures due to compression gradually increases, which slowly affects the bounding box results of grasp detection.}
\label{fig9}
\end{figure*}

\par Cornell datasets are all composed of a single object target, which is less difficult to grasp. To further refine our choice of compression ratio, we performed the same experiment on the multi-object grasp dataset OCID. In the case of fewer objects and fewer stacking occlusions, the grasping detection accuracy does not change too much with the reduction of the compression ratio. However, when the number of objects increases and numerous stacks appear, the influence of different compression ratios on the results gradually appears. As shown in Fig \ref{fig9}, the first line is the result of the grasp detection of the original image, and the eight lines below are the results under the compression ratio of 16.29\%, 4.07\%, 0.99\%, 0.5\%, 0.13\%, 0.05\%, 0.04\%, and 0.03\%(additional images are shown in figure X in the appendix). When the compression ratio is relatively large, the reduction of the compression ratio does not mean the reduction of the accuracy. In some cases, the interference of impurities may even be eliminated to improve the accuracy of grasping detection. However, we can clearly see that when the compression ratio is reduced to 0.5\%, it is difficult to distinguish the stacked occluded objects. When the compression ratio reaches 0.03\% of the limit, it is impossible to perform grasping detection. Therefore, we think that the compression ratio of at least 0.5\% should be selected for multi-object grasp detection.

\begin{figure*}
\setlength{\abovecaptionskip}{-0.75cm}
\begin{center}
\includegraphics[height=7cm]{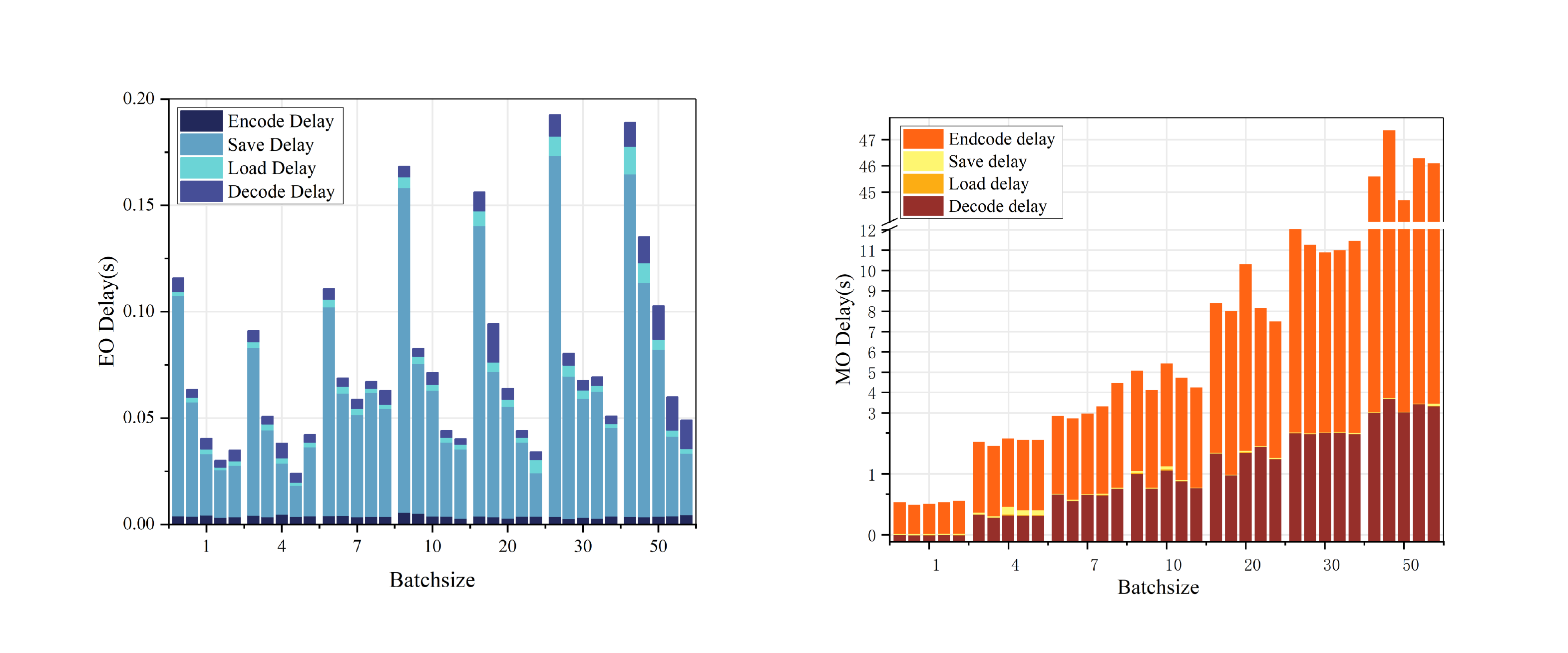}
\end{center}
\caption{Delay under EO and MO. The overall delay consists of four parts: encode delay, save delay, load delay, and decode delay.}
\label{fig10.pdf}
\vspace{-1em} 
\end{figure*}

\begin{figure*}
\setlength{\abovecaptionskip}{-0.75cm}
\begin{center}
\includegraphics[height=5cm]{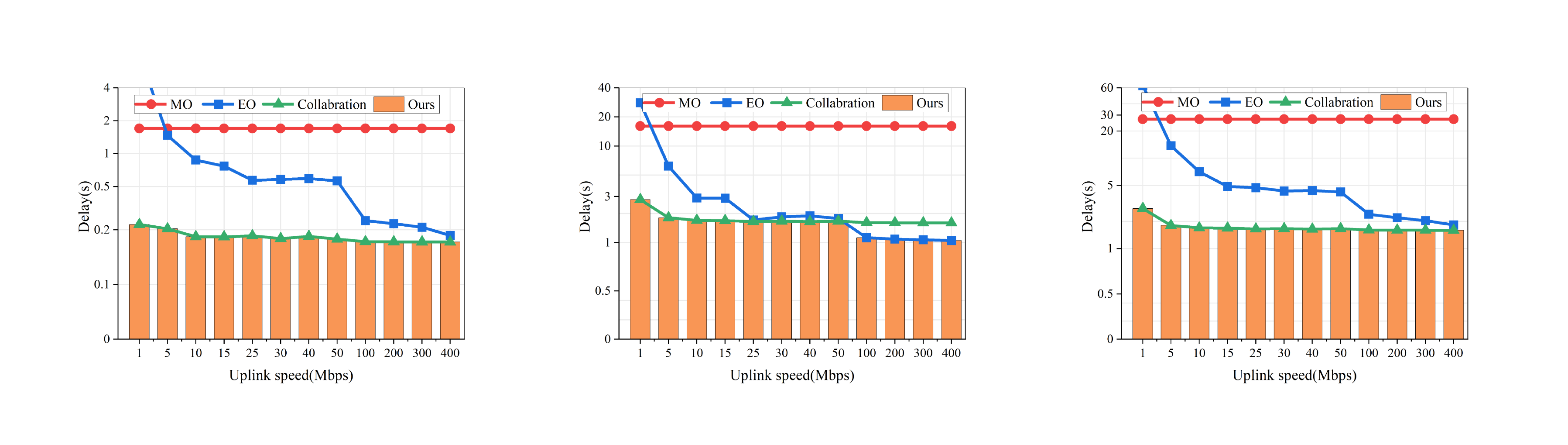}
\end{center}
\caption{Comparison of the overall delay time of four schemes under different network bandwidths. From left to right, the overall delay time for batch sizes of 1,4,10 is shown in sequence.}
\label{fig11.pdf}
\end{figure*}

\par In conclusion, we can draw a conclusion. In the case of a single object or objects without stack occlusion, the compression ratio of 0.13\% or above has high accuracy. In a complex scene where multiple objects or objects are stacked and blocked, a compression ratio of 0.5\% or above is required.

\subsubsection{Changing Uplink Rate Environment Experiment}
\par In practical application situations, the network environment often fluctuates and brings bandwidth changes. With the deterioration of the network environment, the network transmission delay will increase. This makes it necessary for us to choose flexibly among various schemes according to the actual situation. We design experiments to verify how to choose under different network speed conditions.

\par There are several schemes in the experiment:
\begin{itemize}
  \item [a)]
  Pure Edge Offloading (EO): All received pictures will be transmitted to the edge server for calculation and then the data will be returned to the local.
  \item [b)]
  Pure On-device Processing (MO): All the received frames will be directly calculated locally and will not be transmitted to the server.
  \item [c)]
    Collaborative scheme (Collaboration): Preprocess the acquired image locally by encoding and compression, then transmit it to the edge for calculation and finally return to the mobile device.
  \item [d)]
    Our model: According to the real-time uplink network speed, improve the grasp scheme and select the optimal scheme under different network speeds.
\end{itemize}

\par We test the delay on different devices and get Fig \ref{fig10}. The x-axis in the figure represents the number of file frames transmitted in a batch, the compression rate in each group decreases in turn, and the y-axis is the delay time. EO in Fig \ref{fig10}  shows that the time for encoding on the CPU is much less than that for decoding. MO in Fig \ref{fig10} shows that the decoding speed on the GPU is faster so that the reconstruction takes less time than the loading and saving operation of the model. It can be seen that encoding on the CPU and decoding on the GPU are feasible and can make good use of resources. In order to increase efficiency, we explore the impact of the number of frames transmitted at the same time on latency. It is worth noting that the delay of mobile devices fluctuates greatly due to the number of files. The lower compression rate and the smaller tensor volume are beneficial to speed up the loading of the data obtained by the edge server.

\par We chose a fixed compression rate of 2.03\% for transmission experiments under different network bandwidths. Fig \ref{fig11} shows the delay rates of the four schemes under different network bandwidths. It can be seen that the effects of EO and MO vary greatly in different network conditions. The collaboration method can achieve a balanced result between the two. In most cases, the collaboration method can already achieve a good effect and greatly improve the latency of other solutions. However, in some cases, as shown in Fig \ref{fig11}, when the delay of file transfer does not become the key factor, it is possible that EO and MO will achieve better results than the collaboration scheme. So we established a linear model to switch between these three network models, seeking results that are more suitable for multiple factors and situations.

\section{Conclusion}
\par We propose a new grasping detection model and perform grasping detection in RGB images. With the scheme of multi-object multi-grasp, our model improves the mission success ratio of grasping. With the help of edge-cloud collaboration, the computing task is transferred to the cloud with powerful computing power, which greatly improves the speed and accuracy of grasp detection. The encoder and decoder trained by GAN enable the image to be encrypted while compressing, ensuring the security of privacy. The model proves that the combination of autonomous robot grasping and edge-cloud collaboration has great prospects. The model achieves 92\% accuracy on the OCID dataset, the image compression ratio reaches 2.03\%, the structural difference value is higher than 0.91, and the average detection speed reaches 13.62fps. Furthermore, we have packaged our model as a functional package of the ROS operating system, which can be easily used in actual robotic arm operations. In the future, we will improve compression, and refine the distribution of tasks between on-premises and cloud to further improve the efficiency of the model. At the same time, our solution can be fully applied to other work of robots to promote the development of the field of robotics. This work is also potential in some other fields as federated learning \cite{liu2019lifelong, liu2019federated, zheng2021applications}, cloud-edge cooperate robotics \cite{liuelasticros, liu2021peer}, data collection \cite{akcindata}, smart city, etc.

\section{ACKNOWLEDGMENT}
\par This work was supported by Hainan Provincial Natural Science Foundation of China (Grant No. 620MS021), the Key Research and Development Program of Hainan Province (Grant No. ZDYF2021GXJS003, ZDYF2020040), the Major science and technology project of Hainan Province (Grant No. ZDKJ2020012), National Natural Science Foundation of China (NSFC) (Grant No. 62162022, 62162024), the Key Laboratory of PK System Technologies Research of Hainan, Science and Technology Development Center of the Ministry of Education Industry-university-Research Innovation Fund(2021JQR017).

\section{Data Availability}
\par All data included in this study are available from the first author or corresponding author upon reasonable request.

\section{Conflicts of Interest}
\par The authors declare that there are no known conflicts of interest or personal relationships that may affect the work of this research report.

\columnsep 0.12in

\bibliographystyle{IEEEtran}  
\bibliography{my} 

\begin{thebibliography}{10}
\providecommand{\url}[1]{#1}
\csname url@samestyle\endcsname
\providecommand{\newblock}{\relax}
\providecommand{\bibinfo}[2]{#2}
\providecommand{\BIBentrySTDinterwordspacing}{\spaceskip=0pt\relax}
\providecommand{\BIBentryALTinterwordstretchfactor}{4}
\providecommand{\BIBentryALTinterwordspacing}{\spaceskip=\fontdimen2\font plus
\BIBentryALTinterwordstretchfactor\fontdimen3\font minus
  \fontdimen4\font\relax}
\providecommand{\BIBforeignlanguage}[2]{{%
\expandafter\ifx\csname l@#1\endcsname\relax
\typeout{** WARNING: IEEEtran.bst: No hyphenation pattern has been}%
\typeout{** loaded for the language `#1'. Using the pattern for}%
\typeout{** the default language instead.}%
\else
\language=\csname l@#1\endcsname
\fi
#2}}
\providecommand{\BIBdecl}{\relax}
\BIBdecl

\bibitem{sanchez2018robotic}
J.~Sanchez, J.-A. Corrales, B.-C. Bouzgarrou, and Y.~Mezouar, ``Robotic
  manipulation and sensing of deformable objects in domestic and industrial
  applications: a survey,'' \emph{The International Journal of Robotics
  Research}, vol.~37, no.~7, pp. 688--716, 2018.

\bibitem{bicchi2000robotic}
A.~Bicchi and V.~Kumar, ``Robotic grasping and contact: A review,'' in
  \emph{Proceedings 2000 ICRA. Millennium conference. IEEE international
  conference on robotics and automation. Symposia proceedings (Cat. No.
  00CH37065)}, vol.~1.\hskip 1em plus 0.5em minus 0.4em\relax IEEE, 2000, pp.
  348--353.

\bibitem{zhihong2017vision}
C.~Zhihong, Z.~Hebin, W.~Yanbo, L.~Binyan, and L.~Yu, ``A vision-based robotic
  grasping system using deep learning for garbage sorting,'' in \emph{2017 36th
  Chinese control conference (CCC)}.\hskip 1em plus 0.5em minus 0.4em\relax
  IEEE, 2017, pp. 11\,223--11\,226.

\bibitem{kumra2017robotic}
S.~Kumra and C.~Kanan, ``Robotic grasp detection using deep convolutional
  neural networks,'' in \emph{2017 IEEE/RSJ International Conference on
  Intelligent Robots and Systems (IROS)}.\hskip 1em plus 0.5em minus
  0.4em\relax IEEE, 2017, pp. 769--776.

\bibitem{Leoni1998ImplementingRG}
F.~Leoni, M.~Guerrini, C.~Laschi, D.~Taddeucci, P.~Dario, and A.~Starita,
  ``Implementing robotic grasping tasks using a biological approach,''
  \emph{Proceedings. 1998 IEEE International Conference on Robotics and
  Automation (Cat. No.98CH36146)}, vol.~3, pp. 2274--2280 vol.3, 1998.

\bibitem{quillen2018deep}
D.~Quillen, E.~Jang, O.~Nachum, C.~Finn, J.~Ibarz, and S.~Levine, ``Deep
  reinforcement learning for vision-based robotic grasping: A simulated
  comparative evaluation of off-policy methods,'' in \emph{2018 IEEE
  International Conference on Robotics and Automation (ICRA)}.\hskip 1em plus
  0.5em minus 0.4em\relax IEEE, 2018, pp. 6284--6291.

\bibitem{2018Real}
F.~J. Chu, R.~Xu, and V.~Patricio, ``Real-world multiobject, multigrasp
  detection,'' \emph{IEEE Robotics and Automation Letters}, vol.~3, pp.
  3355--3362, 2018.

\bibitem{sarker2019offloading}
V.~K. Sarker, J.~P. Queralta, T.~N. Gia, H.~Tenhunen, and T.~Westerlund,
  ``Offloading slam for indoor mobile robots with edge-fog-cloud computing,''
  in \emph{2019 1st international conference on advances in science,
  engineering and robotics technology (ICASERT)}.\hskip 1em plus 0.5em minus
  0.4em\relax IEEE, 2019, pp. 1--6.

\bibitem{tanwani2020rilaas}
A.~K. Tanwani, R.~Anand, J.~E. Gonzalez, and K.~Goldberg, ``Rilaas: Robot
  inference and learning as a service,'' \emph{IEEE Robotics and Automation
  Letters}, vol.~5, no.~3, pp. 4423--4430, 2020.

\bibitem{deng2016optimal}
R.~Deng, R.~Lu, C.~Lai, T.~H. Luan, and H.~Liang, ``Optimal workload allocation
  in fog-cloud computing toward balanced delay and power consumption,''
  \emph{IEEE internet of things journal}, vol.~3, no.~6, pp. 1171--1181, 2016.

\bibitem{dhawan2011review}
S.~Dhawan, ``A review of image compression and comparison of its algorithms,''
  \emph{International Journal of Electronics \& Communication Technology,
  IJECT}, vol.~2, no.~1, pp. 22--26, 2011.

\bibitem{balle2018variational}
J.~Ball{\'e}, D.~Minnen, S.~Singh, S.~J. Hwang, and N.~Johnston, ``Variational
  image compression with a scale hyperprior,'' \emph{arXiv preprint
  arXiv:1802.01436}, 2018.

\bibitem{Toderici2016VariableRI}
G.~Toderici, S.~M. O'Malley, S.~J. Hwang, D.~Vincent, D.~C. Minnen, S.~Baluja,
  M.~Covell, and R.~Sukthankar, ``Variable rate image compression with
  recurrent neural networks,'' \emph{CoRR}, vol. abs/1511.06085, 2016.

\bibitem{Toderici2017FullRI}
G.~Toderici, D.~Vincent, N.~Johnston, S.~J. Hwang, D.~C. Minnen, J.~Shor, and
  M.~Covell, ``Full resolution image compression with recurrent neural
  networks,'' \emph{2017 IEEE Conference on Computer Vision and Pattern
  Recognition (CVPR)}, pp. 5435--5443, 2017.

\bibitem{Jiang2021OnlineMA}
W.~Jiang, W.~Wang, S.~Li, and S.~Liu, ``Online meta adaptation for
  variable-rate learned image compression,'' \emph{ArXiv}, vol. abs/2111.08256,
  2021.

\bibitem{Chen2020VariableBI}
T.~Chen and Z.~Ma, ``Variable bitrate image compression with quality scaling
  factors,'' \emph{ICASSP 2020 - 2020 IEEE International Conference on
  Acoustics, Speech and Signal Processing (ICASSP)}, pp. 2163--2167, 2020.

\bibitem{Rippel2017RealTimeAI}
O.~Rippel and L.~D. Bourdev, ``Real-time adaptive image compression,'' in
  \emph{ICML}, 2017.

\bibitem{Joshi2021AnalyticalRO}
M.~Joshi, S.~K. Budhani, N.~Tewari, and S.~Prakash, ``Analytical review of data
  security in cloud computing,'' \emph{2021 2nd International Conference on
  Intelligent Engineering and Management (ICIEM)}, pp. 362--366, 2021.

\bibitem{Kaur2015CloudCS}
R.~Kaur and J.~Kaur, ``Cloud computing security issues and its solution: A
  review,'' \emph{2015 2nd International Conference on Computing for
  Sustainable Global Development (INDIACom)}, pp. 1198--1200, 2015.

\bibitem{Li2020AchievingSA}
H.~Li, Y.~Yang, Y.-S. Dai, S.~Yu, and Y.~Xiang, ``Achieving secure and
  efficient dynamic searchable symmetric encryption over medical cloud data,''
  \emph{IEEE Transactions on Cloud Computing}, vol.~8, pp. 484--494, 2020.

\bibitem{Shini2012CloudBM}
S.~Shini, T.~Thomas, and K.~Chithraranjan, ``Cloud based medical image
  exchange-security challenges,'' \emph{Procedia Engineering}, vol.~38, pp.
  3454--3461, 2012.

\bibitem{Yao2014CloudBasedHI}
Q.~Yao, X.~Han, X.~Ma, Y.~Xue, Y.-J. Chen, and J.~song Li, ``Cloud-based
  hospital information system as a service for grassroots healthcare
  institutions,'' \emph{Journal of Medical Systems}, vol.~38, pp. 1--7, 2014.

\bibitem{Radford2016UnsupervisedRL}
A.~Radford, L.~Metz, and S.~Chintala, ``Unsupervised representation learning
  with deep convolutional generative adversarial networks,'' \emph{CoRR}, vol.
  abs/1511.06434, 2016.

\bibitem{He2016DeepRL}
K.~He, X.~Zhang, S.~Ren, and J.~Sun, ``Deep residual learning for image
  recognition,'' \emph{2016 IEEE Conference on Computer Vision and Pattern
  Recognition (CVPR)}, pp. 770--778, 2016.

\bibitem{wang2004image}
Z.~Wang, A.~C. Bovik, H.~R. Sheikh, and E.~P. Simoncelli, ``Image quality
  assessment: from error visibility to structural similarity,'' \emph{IEEE
  transactions on image processing}, vol.~13, no.~4, pp. 600--612, 2004.

\bibitem{Zhao2017LossFF}
H.~Zhao, O.~Gallo, I.~Frosio, and J.~Kautz, ``Loss functions for image
  restoration with neural networks,'' \emph{IEEE Transactions on Computational
  Imaging}, vol.~3, pp. 47--57, 2017.

\bibitem{ren2015faster}
S.~Ren, K.~He, R.~Girshick, and J.~Sun, ``Faster r-cnn: Towards real-time
  object detection with region proposal networks,'' \emph{Advances in neural
  information processing systems}, vol.~28, 2015.

\bibitem{Jia2015GuidingLT}
X.~Jia, E.~Gavves, B.~Fernando, and T.~Tuytelaars, ``Guiding long-short term
  memory for image caption generation,'' \emph{ArXiv}, vol. abs/1509.04942,
  2015.

\bibitem{Agustsson_2017_CVPR_Workshops}
E.~Agustsson and R.~Timofte, ``Ntire 2017 challenge on single image
  super-resolution: Dataset and study,'' in \emph{The IEEE Conference on
  Computer Vision and Pattern Recognition (CVPR) Workshops}, July 2017.

\bibitem{ainetter2021end}
S.~Ainetter and F.~Fraundorfer, ``End-to-end trainable deep neural network for
  robotic grasp detection and semantic segmentation from rgb,'' in \emph{2021
  IEEE International Conference on Robotics and Automation (ICRA)}.\hskip 1em
  plus 0.5em minus 0.4em\relax IEEE, 2021, pp. 13\,452--13\,458.

\bibitem{suchi2019easylabel}
M.~Suchi, T.~Patten, D.~Fischinger, and M.~Vincze, ``Easylabel: a
  semi-automatic pixel-wise object annotation tool for creating robotic rgb-d
  datasets,'' in \emph{2019 International Conference on Robotics and Automation
  (ICRA)}.\hskip 1em plus 0.5em minus 0.4em\relax IEEE, 2019, pp. 6678--6684.

\bibitem{liu2019lifelong}
B.~Liu, L.~Wang, and M.~Liu, ``Lifelong federated reinforcement learning: a
  learning architecture for navigation in cloud robotic systems,'' \emph{IEEE
  Robotics and Automation Letters}, vol.~4, no.~4, pp. 4555--4562, 2019.

\bibitem{liu2019federated}
B.~Liu, L.~Wang, M.~Liu, and C.-Z. Xu, ``Federated imitation learning: A novel
  framework for cloud robotic systems with heterogeneous sensor data,''
  \emph{IEEE Robotics and Automation Letters}, vol.~5, no.~2, pp. 3509--3516,
  2019.

\bibitem{zheng2021applications}
Z.~Zheng, Y.~Zhou, Y.~Sun, Z.~Wang, B.~Liu, and K.~Li, ``Applications of
  federated learning in smart cities: recent advances, taxonomy, and open
  challenges,'' \emph{Connection Science}, pp. 1--28, 2021.

\bibitem{liuelasticros}
\BIBentryALTinterwordspacing
B.~Liu, L.~Wang, and M.~Liu, ``Elasticros: An elastically collaborative robot
  operation system for fog and cloud robotics,'' pp. 1--19, 2022. [Online].
  Available:
  \url{https://www.researchgate.net/publication/363032937_ElasticROS_An_Elastically_Collaborative_Robot_Operation_System_for_Fog_and_Cloud_Robotics}
\BIBentrySTDinterwordspacing

\bibitem{liu2021peer}
B.~Liu, L.~Wang, X.~Chen, L.~Huang, D.~Han, and C.-Z. Xu, ``Peer-assisted
  robotic learning: a data-driven collaborative learning approach for cloud
  robotic systems,'' in \emph{2021 IEEE International Conference on Robotics
  and Automation (ICRA)}.\hskip 1em plus 0.5em minus 0.4em\relax IEEE, 2021,
  pp. 4062--4070.

\bibitem{akcindata}
O.~Akcin, P.-h. Li, S.~Agarwal, and S.~P. Chinchali, ``Data games: A
  game-theoretic approach to swarm robotic data collection,'' in \emph{6th
  Annual Conference on Robot Learning}.

\end{thebibliography}

\clearpage
\begin{appendix} 
\section{Appendix}
\subsection{Grasp detection result under different compression ratios.}

\par
\begin{figure}[!h]
\begin{center}
\includegraphics[height=6.5cm]{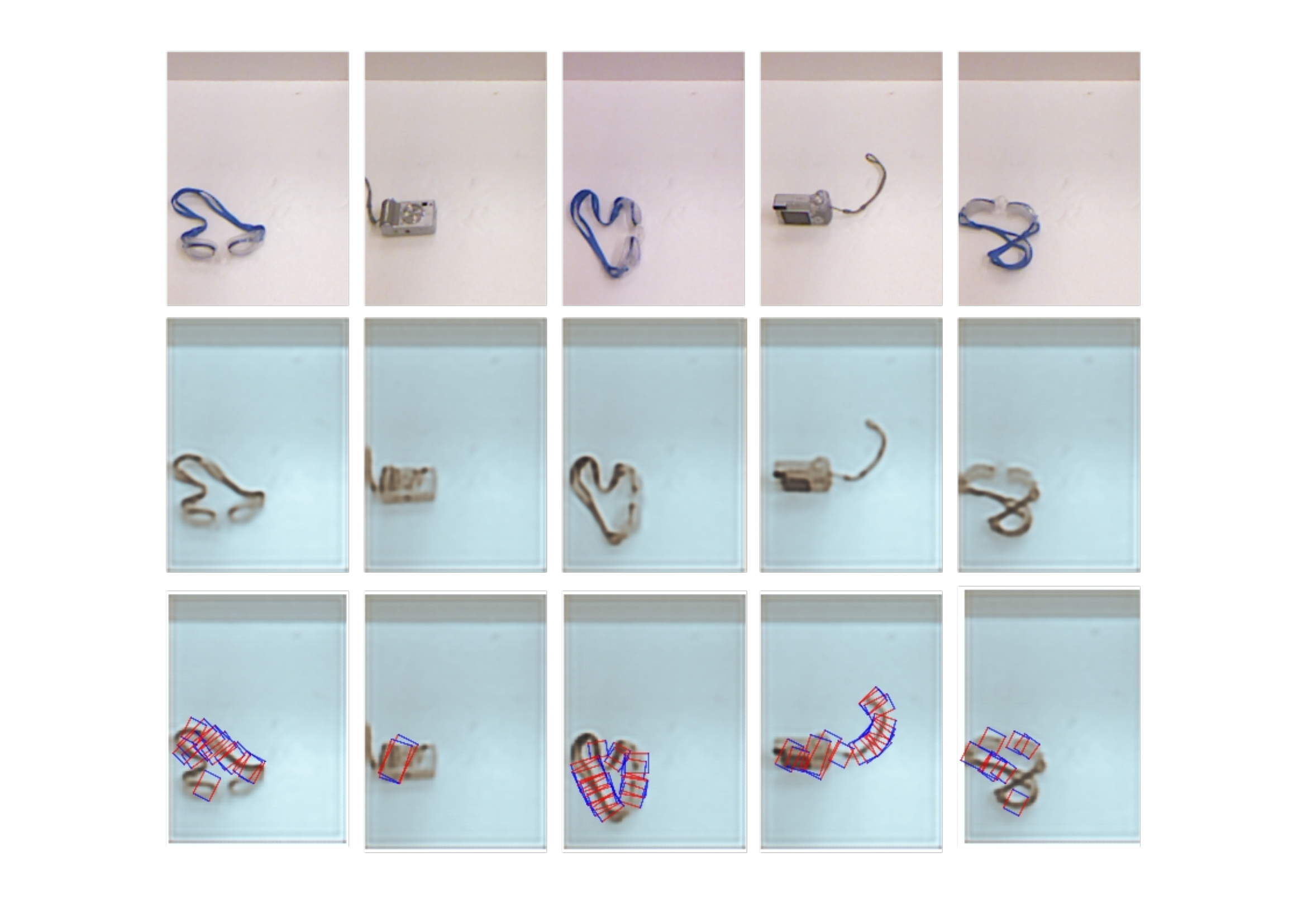}
\end{center}
\caption{Grasp detection result under 0.2\% compression ratio. Under this compression ratio, the grasp detection has high accuracy.}
\label{fig12}
\end{figure}

\begin{figure}[!h]
\begin{center}
\includegraphics[height=6.5cm]{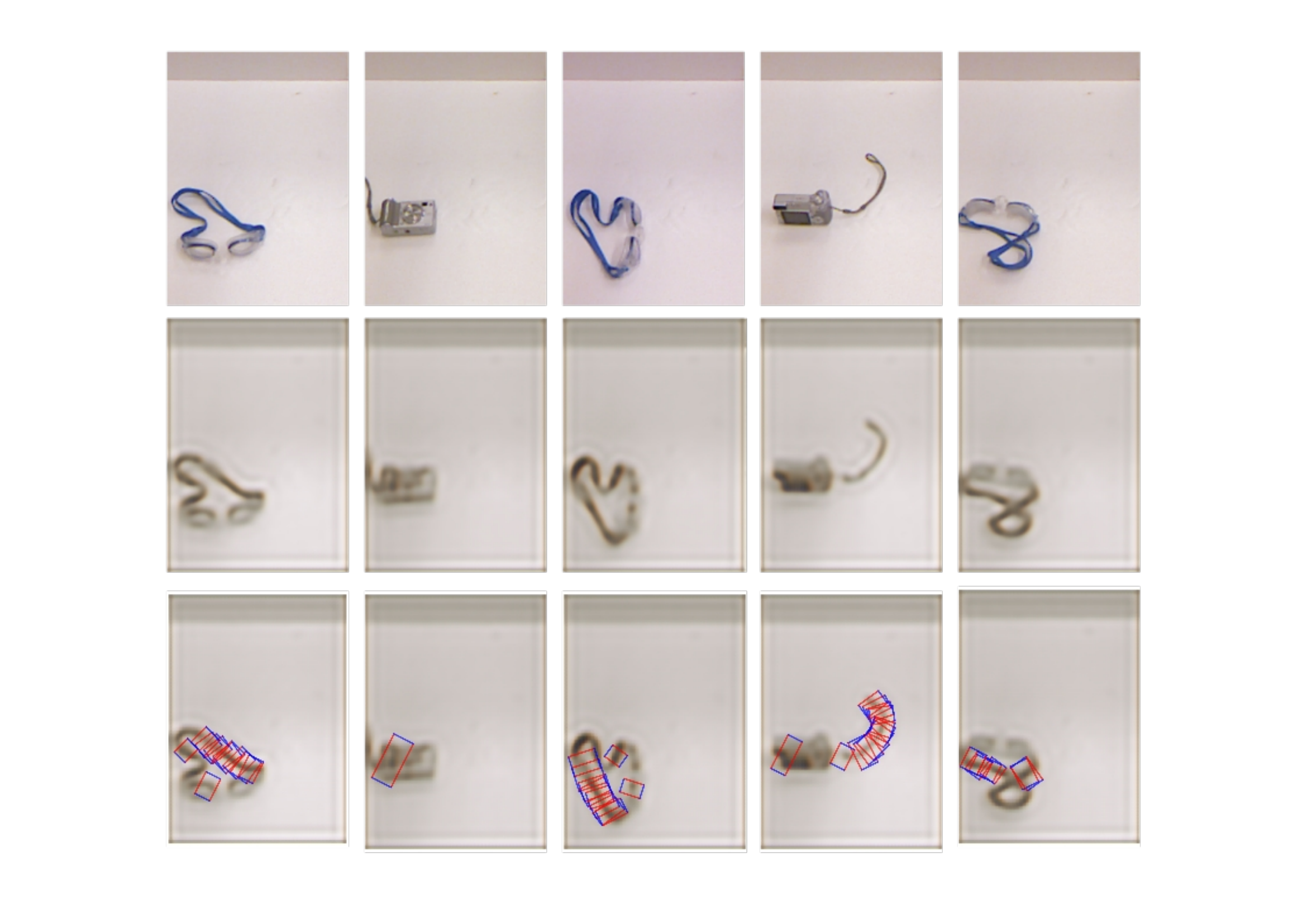}
\end{center}
\caption{Grasp detection result under 0.13\% compression ratio. The image is blurred, but it can still be used for grasp detection.}
\label{fig13}
\end{figure}

\begin{figure}
\begin{center}
\includegraphics[height=6.2cm]{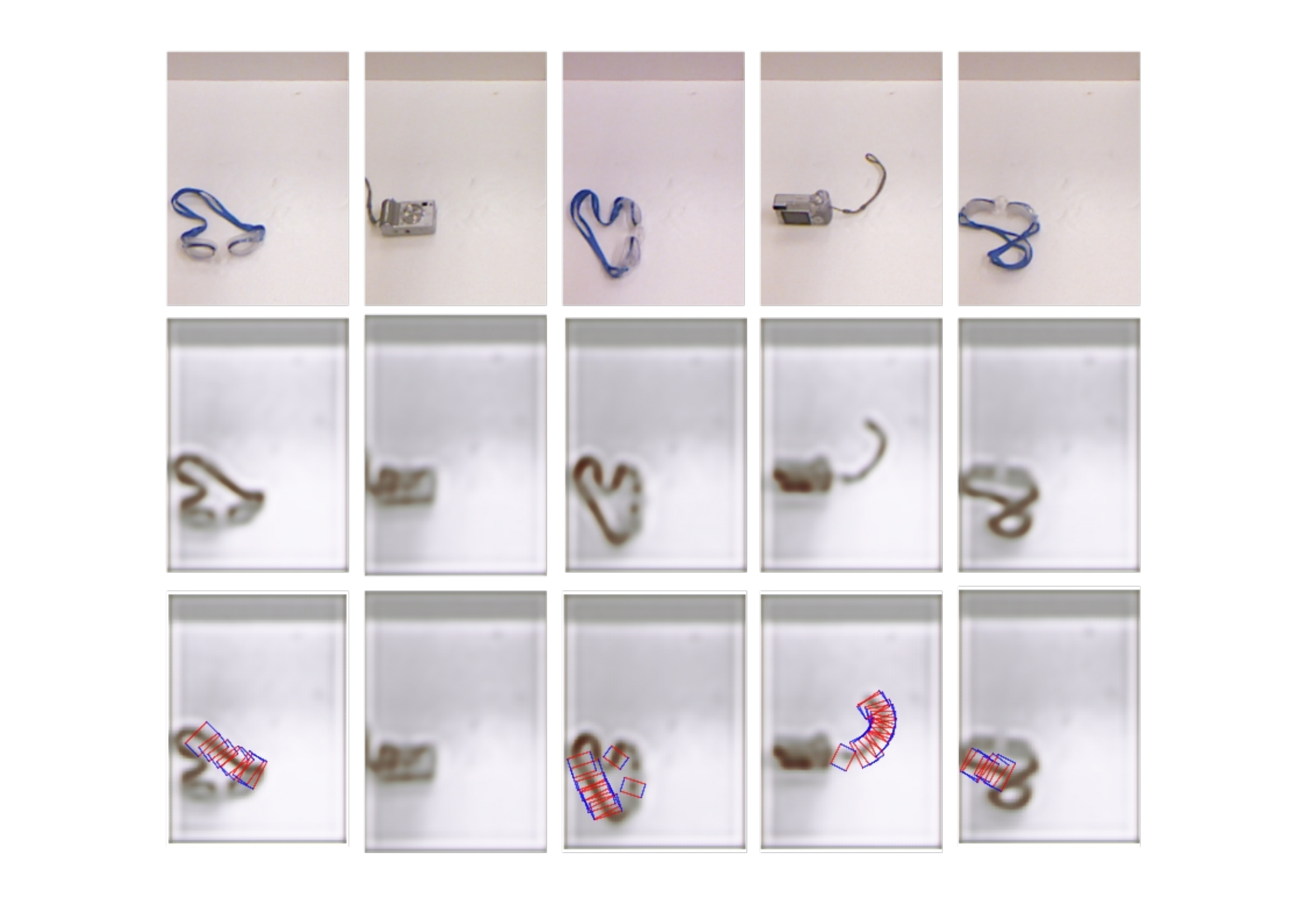}
\end{center}
\caption{Grasp detection result under 0.06\% compression ratio. The loss of image reconstruction becomes large, and some objects cannot be correctly recognized.}
\label{fig14}
\end{figure}

\begin{figure}
\begin{center}
\includegraphics[height=6.2cm]{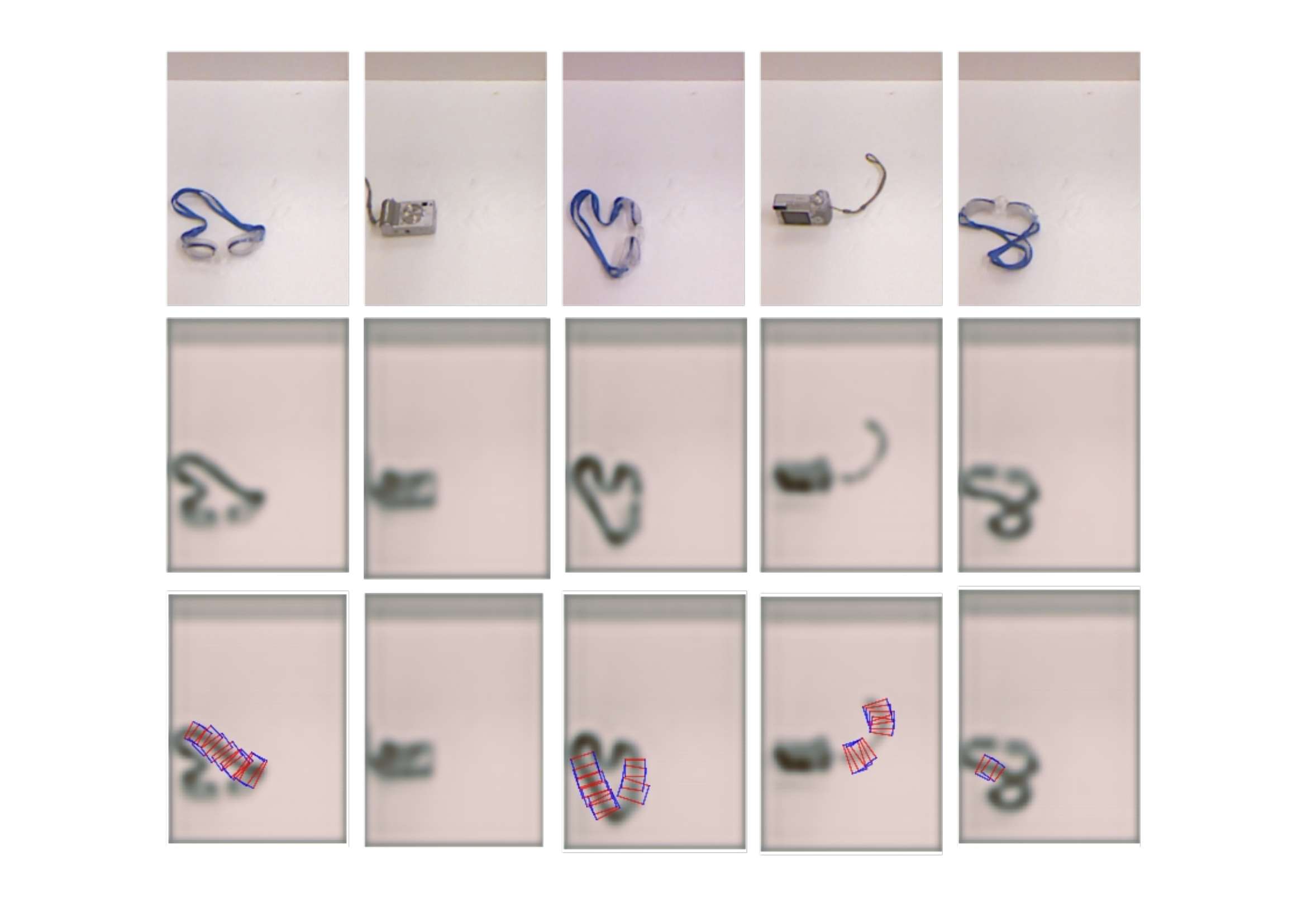}
\end{center}
\caption{Grasp detection result under 0.05\% compression ratio. The image is more blurred, but most objects can still be grasped correctly.}
\label{fig15}
\end{figure}

\begin{figure}[!h]
\begin{center}
\includegraphics[height=6.2cm]{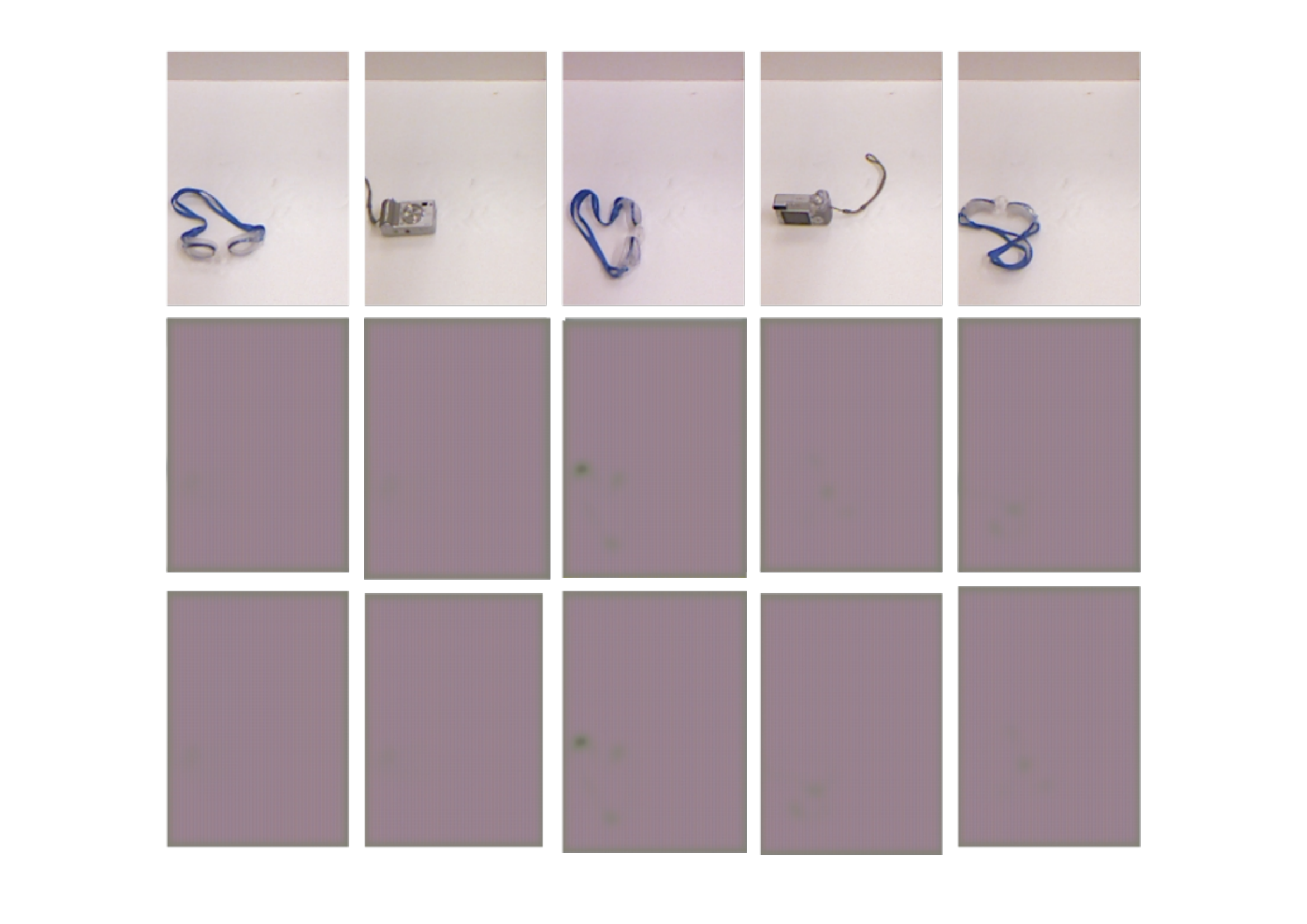}
\end{center}
\caption{Grasp detection result under 0.04\% compression ratio. At this compression ratio, single-object grasp detection cannot be performed.}
\label{fig16}
\end{figure}

\begin{figure*}
\begin{center}
\includegraphics[height=19.5cm]{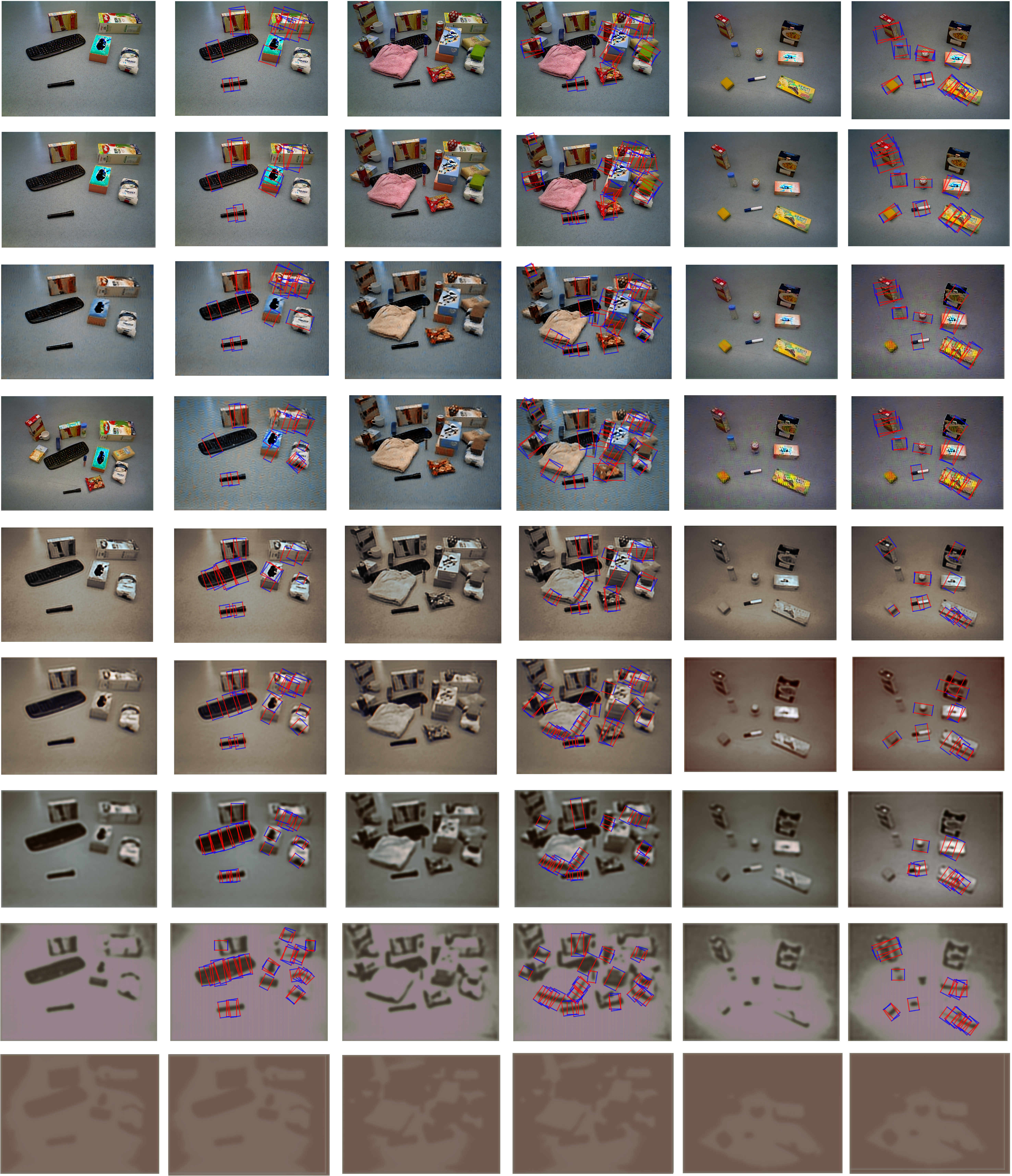}
\end{center}
\caption{The first line is the result of the grasp detection of the original image, and the lines below are the results under the compression ratio of 16.29\%, 4.07\%, 0.99\%, 0.5\%, 0.13\%, 0.05\%, 0.04\%, and 0.03\%. From top to bottom, the loss of pictures due to compression gradually increases, which slowly affects the bounding box results of grasp detection.}
\label{fig17}
\end{figure*}

 The comparison of reconstructed image quality under different channel numbers is shown in Fig.\ref{fig12}-\ref{fig16}. The three lines from top to bottom are the input image, the reconstructed image, and the labeled result. By comparing these five groups of images, it is proved that in a single object grasping task, a low compression ratio can still achieve good results. Until the compression ratio is as low as 0.06\%, it begins to appear that the detected object cannot be recognized and grasped. The results of the multi-object grasp detection task are shown in Fig.\ref{fig17}.

\end{appendix}

\end{document}